# Pan-Arctic Permafrost Landform and Human-built Infrastructure Feature Detection with Vision Transformers and Location Embeddings

Amal S. Perera[1*], *Member IEEE*, David Fernandez[2*], Chandi Witharana[1], Elias Manos[1], Michael Pimenta[1], Anna K. Liljedahl[3], Ingmar Nitze[5], Yili Yang[3], Todd Nicholson[4], Chia-Yu Hsu[6], Wenwen Li[6], Guido Grosse[5].

*Abstract* —Accurate mapping of permafrost landforms, thaw disturbances, and human-built infrastructure at pan-Arctic scale using sub-meter satellite imagery is increasingly critical. Handling petabyte-scale image data requires high-performance computing and robust feature detection models. While convolutional neural network (CNN)-based deep learning approaches are widely used for remote sensing (RS), similar to the success in transformer based large language models, Vision Transformers (ViTs) offer advantages in capturing long-range dependencies and global context via attention mechanisms. ViTs support pretraining via self-supervised learning—addressing the common limitation of labeled data in Arctic feature detection and outperform CNNs on benchmark datasets. Arctic also poses challenges for model generalization, especially when features with the same semantic class exhibit diverse spectral characteristics. To address these issues for Arctic feature detection, we integrate geospatial location embeddings into ViTs to improve adaptation across regions. This work investigates: (1) the suitability of pre-trained ViTs as feature extractors for high-resolution Arctic remote sensing tasks, and (2) the benefit of combining image and location embeddings. Using previously published datasets for Arctic feature detection, we evaluate our models on three tasks—detecting ice-wedge polygons (IWP), retrogressive thaw slumps (RTS), and human-built infrastructure. We empirically explore multiple configurations to fuse image embeddings and location embeddings. Results show that ViTs with location embeddings outperform prior CNN-based models on two of the three tasks including F1 score increase from 0.84 to 0.92 for RTS detection, demonstrating the potential of transformer-based models with spatial awareness for Arctic RS applications.

This paragraph of the first footnote will contain the date on which you submitted your paper for review, which is populated by IEEE. This research was supported by the Google.org's Impact Challenge on Climate Innovation grant, U.S. National Science Foundation (NSF) grant #: 1720875, 1722572, 1927872, 1927723, 1927729. eXtreme Science and Engineering Discovery Environment (Award #: DPP 190001) and Texas Advanced Computing Center Award #: DPP20001) Authors would like to thank Polar Geospatial Center, University of Minnesota for imagery support under NSF-OPP awards 1043681 and 1559691 (Corresponding author Amal S. Perera).

* Equal contribution first authors
[1] Department of Natural Resources and the Environment, University of Connecticut, Storrs, CT. amal.perera, chandi.witharana, elias.manos, michael.pimenta@uconn.edu.
[2] Google, Mountain View, CA, djfernandez@google.com
[3] Woodwell Climate Research Center, Falmouth, MA., aliljedahl@woodwellclimate.org, yyang@woodwellclimate.org
[4] National Center for Supercomputing Applications, University of Illinois, Urbana-Champaign, IL, tcnichol@illinois.edu
[5] Alfred Wegener Institute, Helmholtz Centre for Polar and Marine Research, Potsdam, Germany, ingmar.nitze, Guido.Grosse@awi.de
[6] School of Geographical Sciences and Urban Planning, Arizona State University, Tempe, AZ, chsu53@asu.edu, wenwen@asu.edu.

## I. INTRODUCTION

The demands for advanced monitoring tools to accurately map Arctic permafrost landforms, ice-rich permafrost, track thaw disturbances over time, and assess economic impact of permafrost hazards on human-built infrastructure are rapidly growing [1], [2], [3]. The Arctic is warming up to four times faster than the global average [4], resulting in vulnerabilities, such as enhanced terrestrial [5] and coastal erosion [6], substantial risks to infrastructure [7] and to its large soil carbon pool [8]. The mere presence of ice-rich permafrost as permafrost, a subsurface phenomenon defined by ground temperatures of 0°C or below for at least two consecutive years, cannot be directly measured from space, monitoring the landforms and landscape changes associated with permafrost thaw and ground ice melt is crucial to quantify and predict the evolution of the arctic permafrost region [9]. The entire Arctic has been repeatedly imaged by commercial very high spatial resolution (VHSR) earth observation (EO) satellite sensors, such as Maxar and Planet, enabling the capture of fine-scale changes in microtopographic features [10], [11], [12], thaw disturbances [13], [14], [15], and infrastructure [16], [17]; without compromising spatial detail or geographic coverage. However, despite the rapid growth of VHSR satellite repositories into the petabyte scale, geospatial products derived from this imagery at a pan-Arctic scale remain scarce due to data access, data handling, and analysis challenges. These big data challenges combined with the growing quest for pan-Arctic scale permafrost modeling efforts naturally set the stage for artificial intelligence (AI) methods, such as deep learning (DL) algorithms. Despite the remarkable performances of DL models in everyday image understanding and computer vision (CV) applications, bottlenecks still exist when translated to geospatially explicit EO image analysis tasks. Image characteristics, seasonality, landscape heterogeneity, and most importantly semantic complexity aggregated into multiple spatial scales pose greater friction on the strengths of DL model inferences. The scalability of automated analysis over millions of square kilometers comprising heterogeneous landscapes reverberates the need for efficient AI-enabled image-to-assessment workflows that center on high performance computing resources.

Vision Transformers (ViTs) utilize self-attention to capture both local and global dependencies [18], contrasting with the receptive field limitations of Convolutional Neural Networks (CNNs). ViTs allow each image patch to consider all others, establishing long-range relationships without relying solely on stacked convolutional layers. The dynamic allocation of attention, inherent in self-attention, mirrors aspects of human visual system (HVS) processing, specifically the integration of detailed and contextual information. This enables ViTs to adapt to image content, effectively weighing the importance of patches based on learned features and context, akin to top-down and bottom-up processing in the HVS. Studies have demonstrated ViT's superior performance over CNNs in various vision tasks, attributed to their ability to maintain a global receptive field while preserving spatial relationships [18]. This capability, analogous to the simultaneous handling of spatial scales in human vision [19], facilitates typical CV tasks of image classification, object detection, and segmentation. ViT's ability to maintain a global receptive field, while preserving spatial relationships, is specifically relevant in remote sensing where understanding and utilization of spatial context and relationships is crucial for accurate detection and segmentation of feature classes or landforms. With ViTs, spatial context helps differentiate similar features by considering their relative spatial position in the scene. Knowing that a pixel represents water would not be enough to determine whether it belongs to a lake or an ocean; understanding its spatial context and relationships is essential for that classification. ViTs through the use of their self-attention mechanism are able to better capture these complex spatial relationships allowing for the ability to discern features using both pixel value and spatial relationships in the landscape.

Due to recent advancements in self-supervised learning (SSL) and Masked Autoencoder (MAE) models, ViTs have achieved remarkable success in CV and remote sensing [20] without relying on large troves of labeled data. MAEs are specifically designed to produce task-agnostic representations by learning from large-scale, unlabeled datasets. Once pre-trained, they can be fine-tuned with supervised learning to adapt ViTs for specific tasks. A significant limitation of CNN-based backbones is their dependence on publicly available labeled datasets, which are often not tailored to the unique characteristics or requirements of geospatial EO datasets and even more so for particular sub-domains, such as imagery of landscape dynamics. This mismatch can hinder the model's performance in domain-specific scenarios. In contrast, MAE-based approaches leverage the abundance of high-resolution satellite imagery in remote sensing, enabling the model to extract fine-grained spatial details and global semantic structures [21]. This ability to capture both local and global context is critical for complex remote sensing applications. Although we do not use MAE in this work, its demonstrated success serves as a key inspiration for adopting ViTs. By harnessing the inherent strengths of ViTs, we explore their potential in addressing challenges specific to remote sensing without the reliance on large-scale labeled datasets typical of CNN-based methods.

In CV-based remote sensing, distinguishing features that share the same semantic concept but differ in spectral characteristics across spatial locations poses a significant challenge in model generalization with limited labeled datasets. Location embeddings address this by leveraging geospatial context, incorporating spatial priors that enhance image understanding and feature representation. By integrating geospatial metadata, such as coordinates or regional characteristics, models can disambiguate semantically similar features in different environmental contexts (e.g., urban areas vs. agricultural fields) [22], [23]. Recent advancements, such as GeoCLIP [24], use contrastive learning to align satellite imagery with geographic metadata, capturing rich, location-aware representations [25]. These approaches improve downstream tasks like object detection by embedding regional context and spatial relationships, enhancing generalization across geographies and supporting zero-shot learning for unseen regions. By bridging geographic and visual information, location embeddings offer scalable solutions to key challenges in remote sensing [26], [27].

Considering the potential of ViTs to capture the spatial context and relationships via attention mechanisms, and the ability of deep location embedding models to enable geographic location generalization, in this work we explore the effectiveness of ViTs with location embeddings for remote sensing Arctic features. We systematically explore and obtain a quantitative justification for the potential benefits of leveraging (1) Pre-trained ViTs as the backbone for feature extraction compared to CNNs, and (2) adapt a pre-trained deep location embedding model to enhance model generalization in tasks like semantic and instance segmentation for the pan-Arctic region.

## II. Background
### A. Computer Vision in Remote Sensing

Traditional RS image classification methods, k-nearest-neighbors ($k$-nn), minimum distance classifiers, random forest, and logistic regression [28] primarily relied on spectral features but lacked capacity to model spatial dependencies. Advanced methods, such as spectral-spatial classification [29] and local Fisher discriminant analysis [30], improved feature extraction techniques integrating spatial information. Support Vector Machines (SVMs) emerged as a robust classifier for image analysis and were particularly effective for small training datasets [31] [32]. Support Vector Machines (SVMs) ability to learn an optimal hyperplane for class separation, and extensions to SVMs further enhanced remote sensing tasks [29],[33].

Exploration of Neural networks (NNs), such as multilayer perceptrons (MLPs) [34] and radial basis function (RBF) networks [35], for RS classification were initially constrained by shallow architectures and computational limitations. With deeper architectures, deep learning [36], transformed this landscape, surpassing, SVMs enabling deep learning CV techniques to transition into RS applications. This allowed RS to leverage high-resolution geospatial data for tasks, such as land use classification, environmental monitoring, and disaster response [37], [38], [39]. Modern techniques like fully convolutional networks [40] and transformer-based models [18], [41], are driving continued innovation in geospatial analysis for RS.

A fundamental CV task, semantic segmentation involves assigning a meaningful label to every pixel in an image providing a fine-grained understanding of the image content. This dense prediction task results in a pixel-wise segmentation mask, where each pixel is assigned a class label, such as "sky," "road," "person," or "house". In remote sensing, semantic segmentation is used to analyze satellite images to classify land cover, track deforestation, monitor urban development, or assess natural disasters. Semantic segmentation has progressed from early hand-crafted feature methods with textures, edges, and colors [42], to Conditional Random Fields (CRFs) and Markov Random Fields (MRF) to the use of fully convolutional networks [40] that enabled end to end learning. Further advancements included encoder-decoder architectures [43] and global context modeling using attention and ViTs [41], driving the field forward.

Instance segmentation extends semantic segmentation by not only classifying each pixel but also distinguishing individual instances within the same class—for example, labeling each house separately in a satellite image. This granularity enables richer analysis, especially valuable in remote sensing applications like urban planning, environmental monitoring, and disaster response. The field has progressed from early region-based methods with hand-crafted features to deep learning models, such as MaskRCNN [44], which introduced pixel-level mask predictions. Subsequent advancements include anchor-free methods [45], bottom-up approaches, such as CenterMask [46], and transformer-based models like Mask2Former [47], focusing on improved object relationship modeling via attention mechanisms.

### B. Vision Transformers

A ViT is a transformer model specifically designed for CV tasks. Unlike CNNs, ViTs decompose an image tile into a series of patches, treating them as tokens similar to words in natural language processing (NLP). These patches are serialized into a sequence of vectors and mapped into a higher-dimensional representation, that is processed by a transformer encoder. ViTs introduced by [18], where they demonstrated that pure transformers, originally developed for NLP, could achieve competitive results in vision tasks when trained on large-scale datasets. Following this, the Swin Transformer [48] introduced a hierarchical transformer architecture with shifted windows, improving computational efficiency and scalability for high-resolution images. Further advancements, such as DeiT [49], demonstrated that data-efficient training strategies could make ViTs viable without massive datasets. Figure 1 illustrates how a ViT processes an image by dividing it into patches and applying a transformer encoder to perform downstream vision tasks with a supporting detection head trained for a downstream task.

### C. Feature Detection Architectures

Pioneered by R-CNN [50], object detection, adopted pre-training and fine-tuning for downstream tasks. A general-purpose, task-agnostic backbone network is pre-trained on large-scale datasets, such as ImageNet [51] using supervised learning enabling the model to learn transferable generic, low-level visual features like edges, textures, and shapes. The backbone network, typically a deep convolutional neural network (CNN), such as AlexNet [52], VGG [53], or ResNet [54], is then fine-tuned for a specific downstream task by adding specialized heads for classification, bounding box regression, or segmentation. This approach dramatically reduces the computational cost and time required to train a model from scratch for each new task. Generally the backbone is the core feature extractor in an encoder-decoder object detection model. It processes raw input images and generates multi-scale feature maps that serve as the foundation for downstream tasks. Early backbones include CNN architectures like ResNet [54], which extract hierarchical features from low-level textures to high-level semantic representations. More recently, transformer-based backbones are used [18], [48], [55]. Regardless of architecture, the backbone's pre-training, either supervised (e.g., on ImageNet) or self-supervised, is crucial for initializing the model with transferable features.

The neck bridges the backbone and different types of detection heads, refining and aggregating features across multiple scales. This is particularly important for detecting objects of varying sizes. A popular neck design is the Feature Pyramid Networks (FPNs). These construct a top-down pathway that fuses low-resolution, semantically rich features with high-resolution, spatially detailed features. This pyramid of features is then passed into the detection head which produces the final predictions, including object classifications, bounding box coordinates, and optionally, masks for instance segmentation.

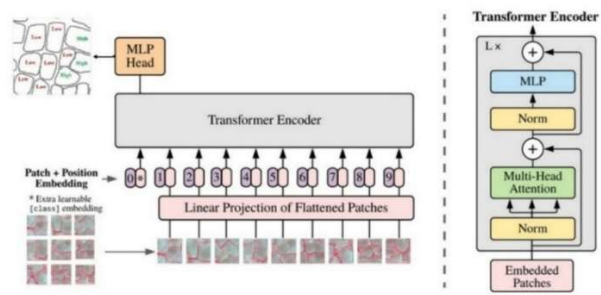

**Figure 1.** A possible application of a ViT for remote sensing a downstream task based on the model introduced by [18]. Each image tile is patched and combined with a positional embedding before being consumed by the transformer block. Features extracted from the Transformer are used by the downstream detection head to complete the detection task. The transformer encoder consists of multiple layers (L) of transformers.

### D. Location Embedding in Remote Sensing

Location embeddings have emerged as a powerful tool in RS to incorporate geospatial context into models, addressing challenges, such as distinguishing features with similar spectral signatures in varying geographic settings. Early efforts leveraged geographic metadata, such as coordinates or regional labels, to enrich feature representations. For example, [22] demonstrated the utility of geospatial priors for disambiguating visually similar features across different environmental contexts. Building on these concepts, subsequent research explored integrating geospatial information into deep learning models. [23] highlighted the potential of using auxiliary spatial metadata alongside image data for improved segmentation and classification tasks in remote sensing. Contrastive learning frameworks like CLIP [25] extended this approach by aligning visual features with textual or geographic descriptors, creating rich, location-aware embeddings that improved model generalization. Recent advancements have further refined these methodologies to geospatial data. GeoCLIP [24] adapted the CLIP framework to encode meaningful correspondences between satellite imagery and geographic metadata, such as land cover labels and textual descriptions. These embeddings have proven effective in enabling models to generalize across diverse geographies and achieve zero-shot learning, where unseen regions can be analyzed without extensive retraining.

[27] introduced a novel approach to location embeddings that stands out from prior work by focusing on explicitly modeling spatial hierarchies and regional dependencies within geospatial data. Unlike earlier methods, which often relied on straightforward integration of geographic metadata (e.g., coordinates or region labels) into feature representations, this work proposed a hierarchical location embedding framework. SatCLIP introduced a multi-scale encoding approach that represents geographic features at various spatial resolutions [27]. This ensures that models can effectively distinguish between fine-grained, local-level variations (e.g., differences between adjacent urban blocks) and broader regional patterns (e.g., climatic or ecological zones). Unlike static embeddings, their method dynamically contextualizes geospatial data based on its surrounding environment, allowing the embeddings to adapt to varying spatial relationships across different geographic settings. This makes the embeddings more robust when applied to regions outside the training dataset. The work by Klemmer and others extensively evaluated hierarchical location embedding framework across highly diverse geographic regions, showcasing the method's generalization capabilities. The embeddings outperformed existing approaches in tasks where geospatial context played a critical role

## III. METHOD

There are multiple possible architectural considerations when adopting and combining vision transformers and location embeddings. We adopted an empirical approach for the models' architectural design decisions. We relied on systematic experimentation and observations to guide decisions about model structure and configurations. This type of empirical approach can be particularly important in deep learning because the behavior of models often depends on nuanced interactions between data, architecture, and optimization processes, which cannot always be derived from theory. It also enables the discovery of effective and sometimes unexpected solutions tailored to specific tasks or datasets.

### A. Proposed Architecture

We will first describe how we leverage the ViTDet architecture introduced by [56] as a way of producing a pyramidal feature structure that can then be consumed by a task specific head. Below we provide a brief overview of the ViTDet architecture. The backbone of a ViTDet feature extractor is a plain ViT which follows the typical pattern of patchifying the input image, embedding the patches through a linear projection, and adding a positional encoding to preserve spatial information. These enriched embeddings are subsequently processed by the ViT backbone encoder, which consists of multiple transformer layers. To help reduce computational complexity and training cost without compromising too much on the benefits of global attention, the ViTDet work leverages a combination of window and global attention. They split the backbone into subsets of blocks where only the last block in each subset performs global attention. The output from the final block of the ViT backbone is reshaped from a set of flattened patches into a spatial grid to reconstruct the image-like structure [56]. The spatially structured feature map $F$ is passed through a simple feature pyramid network (SFPN):

$$P_i = SFPN(F, S_f, S_d, C_d) \in \mathbb{R}^{C_d \times H_i \times W_i}$$
$$for\ i \in \{16, 8, 4, 2\} \quad (1)$$

where $P_i$ represents feature maps at different pyramid levels, $S_f = 16$ represents the scale of F, $S_d = i$ represents the desired scale of each $P_i$, $C_d = 256$ represents the desired channels of each $P_i$, and $H_i = \frac{H}{S_d}$ and $W_i = \frac{W}{S_d}$ represent the height and width of each $P_i$. To produce each $P_i$ Algorithm 1 below is run in parallel for each level of the pyramid:

```
Algorithm 1 Simple Feature Pyramid Network
 1: Inputs: ViT feature map F w/ shape (C, H/S_f, W/S_f), feature map
    scale S_f, desired scale S_d, desired channels C_d
 2: num_resizes := log_2(S_f) - log_2(S_d)
 3: for i ∈ range(|num_resizes|) do
 4:     if num_resizes < 0 then
 5:         // This means we need to down sample
 6:         F := MaxPool2D(F)
 7:     else
 8:         // This means we need to up sample
 9:         if i > 0 then
10:             F := GeLU(LayerNorm(F))
11:         end if
12:         F := Deconvolution2D(F)
13:     end if
14: end for
15: F := LayerNorm(Conv2D_{1x1}(F, C_d))
16: F := LayerNorm(Conv2D_{3x3}(F, C_d))
17: return F // Shape (C_d, H/S_d, W/S_d)
```

The resulting structure allows us to construct a multi-scale feature map from the single resolution output of the ViT backbone without the need for complex hierarchical structures [56]. The global self-attention mechanism in ViTs allows each feature to inherently encode information from the entire image, reducing the reliance on explicit hierarchical feature processing for scale information. Simple feature pyramids also reduce architectural complexity, as they avoid the need for intricate multi-resolution designs. This is computationally efficient and easier to integrate with transformer-based backbones. The authors demonstrate that a plain ViT backbone paired with a simple feature pyramid can match or surpass the performance of traditional hierarchical designs in object detection tasks, suggesting that hierarchical structures are not strictly necessary for transformer-based models [56]. In the following subsections, we explore how we combined a ViTDet feature extractor with task specific heads to solve instance and semantic segmentation problems.

*Semantic Segmentation.*

For semantic segmentation, we chose to build on the TransUNet work proposed by [57] in which they explore adapting a CNN based UNet to leverage a ViT backbone for medical image segmentation. To circumvent the lack of hierarchical features in a plain ViT, they propose using a ResNet-50 [54] as a way of encoding the input image into a feature map and passing it into a ViT encoder to produce a final representation [57]. This allowed them to leverage the intermediate representations produced by the ResNet-50 encoder and the final output of the ViT for the skip connections typically found in UNet-style architectures. To upsample the final representation produced by the ViT, they introduce a cascaded upsampler which consists of multiple upsampling blocks each comprised of a 2× bilinear upsampling operator and a 3×3 convolution layer followed by a ReLU non-linearity [57]. Once the features are upsampled to the original $H \times W$, a $1 \times 1$ convolution is applied to the feature map to produce a set of logits Y with shape $N \times H \times W$ where N is the number of ground truth classes. The logits Y are converted into class probabilities via a softmax function, and the final semantic segmentation map is obtained by taking the argmax over the class dimension. The left-hand side of Figure 2 below provides a visual representation of their work.

For our semantic segmentation architecture, we modify the TransUNet architecture by replacing the CNN encoder with a SFPN. We do this by leveraging the multi-scale feature map produced by the SFPN to construct the cross connections used by the cascaded upsampler to help with localization of the upsampled feature map. This removes the need for pre-training a CNN encoder and provides all of the benefits mentioned above from using a ViTDet feature extractor. A visualization of our adapted architecture is shown on the right-hand side of Figure 2 below, we call it ViTDet-UNet.

*Instance Segmentation.*

For the instance segmentation task, we directly use the ViTDet work in which they adapt the Mask-RCNN framework to work with the outputs of the simple feature pyramid [56]. The SFPN features $P_i$ are passed through a Region Proposal Network (RPN) to generate proposals for object instances. The RPN predicts bounding boxes and for each object proposal generated by the Region Proposal Network (RPN), the Region of Interest (ROI) Align operation is performed. ROI Align extracts fixed-size feature maps from the feature pyramid. These extracted features are then fed into two parallel heads: a box head and a mask head. The box head processes the extracted features to predict the class and bounding box of the object instance. It outputs class probabilities enabling multi-class instance segmentation and the bounding box coordinates represented by four values ($X$, $Y$, $W$, $H$). Simultaneously, the mask head takes the same extracted features and generates pixel-wise logits, representing the likelihood of each pixel belonging to the object instance. A sigmoid activation function is applied to the pixel-wise logits generated by the mask head, producing probability maps. Each probability map represents a binary mask, indicating the probability that a given pixel is part of the object instance. These probability maps are subsequently thresholded to create the final binary masks. In this work we do not introduce this ViT-based variation to Mask-RCNN but for the purpose of discussion, we name it ViTDet-Mask-RCNN. Figure 3 below visualizes the distinctions between the original and the adaptation presented by [56].

### B. Location Embeddings with SatCLIP

Detecting features in the Arctic presents unique challenges due to its harsh environmental conditions (snow/ice, cloudiness, low sun angles), dynamic and diverse landscapes, and limited labeled data. SatCLIP, a contrastive learning-based approach tailored for satellite imagery, offers a powerful solution by leveraging pre-trained models that align geospatial information and visual representations. This enables rich feature extraction even in data-scarce regions. We integrate location embeddings using SatCLIP to both semantic and instance segmentation models to enhance their spatial awareness. The multiple architecture variants when selecting the encoder size, placement and the merging strategy are briefly discussed below.

*Encoders*

The number of Legendre polynomials (L) used in the model's spherical harmonics corresponds to the resolution at which the SatCLIP models are trained and determines its sensitivity. Smaller values of **L** prioritize computational efficiency and are well-suited for capturing broad, large-scale spatial patterns, while larger values enable finer spatial distinctions. **SatCLIP L10** and **SatCLIP L40** represent two variants optimized for different spatial resolutions. **L10**, with a lower-order location encoding, is expected to generalize well across broad geographic regions, making it more suitable for tasks requiring large-scale spatial reasoning. In contrast, **L40**, with a higher-order encoding, provides more detailed spatial representations, which may be advantageous for tasks requiring localized spatial precision, such as regional analyses or interpolation. However, higher-resolution models may also be more sensitive to dataset biases and overfitting [27].

*Encoder Placement*

Location embedding integration can be achieved by incorporating location embeddings at two different stages in the prediction pipeline architecture (Figure 4), such as prior to the Simple Feature Pyramid (SFP) and post-SFP. Pre-SFP would enable the model to incorporate location embeddings as part of the up/down sampling process, fusing the location information into the various levels of resolution. On the other hand, the post-SFP merging approach would provide the raw location representation at each scale.

*Merge Strategy*

Either adding or concatenating can be used for integrating the two encodings. With adding, the location embeddings are elementwise added to the feature embeddings. This technique operates as a form of positional bias. It enhances the features by directly integrating spatial information without increasing the dimensionality. With concatenating, the location embeddings are concatenated along the feature dimension with the image embeddings. This expands the dimensionality of the feature representation, allowing the model to treat visual and spatial information as separate but equally important channels allowing the model to independently learn the importance of visual and location features. When employing either of these techniques, one also has to consider whether to normalize the location embeddings or to normalize the combined embeddings. For feature embeddings F and location embeddings L, there are multiple merging strategies to produce the combined embeddings C as given in equations (2 - 9) where $\frac{F}{|F|}$ indicates normalization, + indicates adding, ***Concat*** indicates concatenation of embeddings:

$$C = F + L \qquad (2)$$

$$C = Concat(F, L) \qquad (3)$$

$$C = F' + L', \text{ where } F' = \frac{F}{|F|} \text{ and } L' = \frac{L}{|L|} \qquad (4)$$

$$C = Concat(F', L') \text{ where } F' = \frac{F}{|F|} \text{ and } L' = \frac{L}{|L|} \qquad (5)$$

$$C = \frac{C'}{|C'|} \text{ where } C' = Concat(F, L) \qquad (6)$$

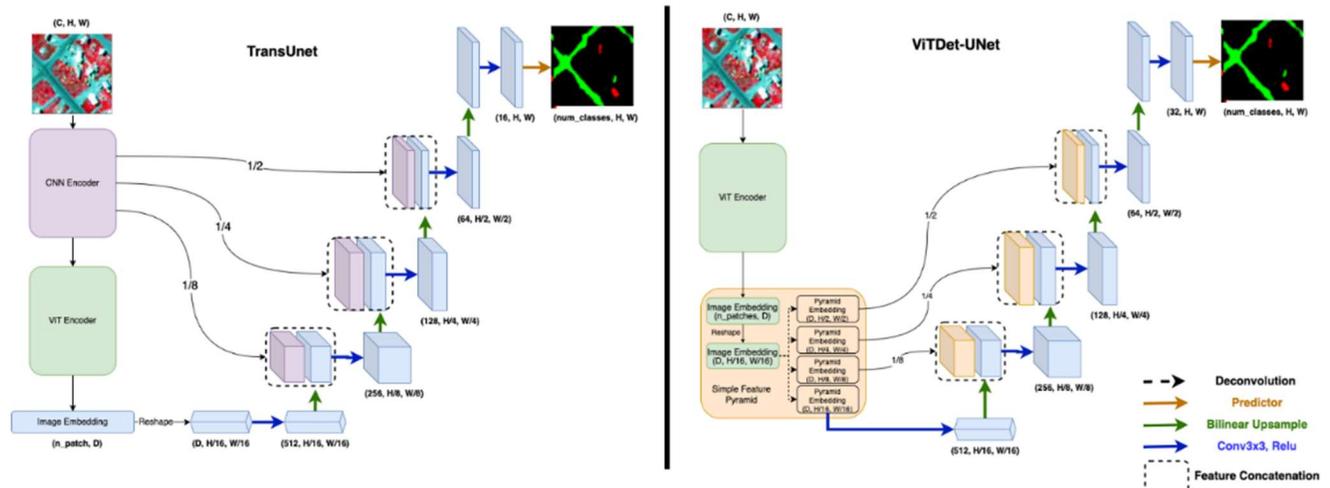

**Figure 2.** Left hand side represents the original TransUNet architecture proposed by Chen et al. The right-hand side shows our proposed ViTDet-UNet architecture which substitutes the CNN encoder for a feature pyramid in order to create the hierarchical feature maps.

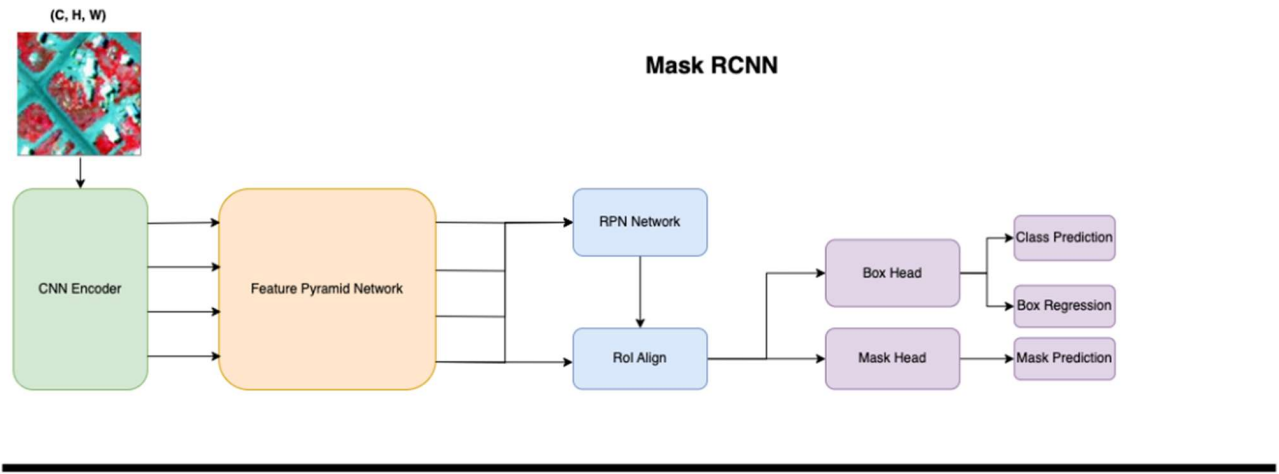

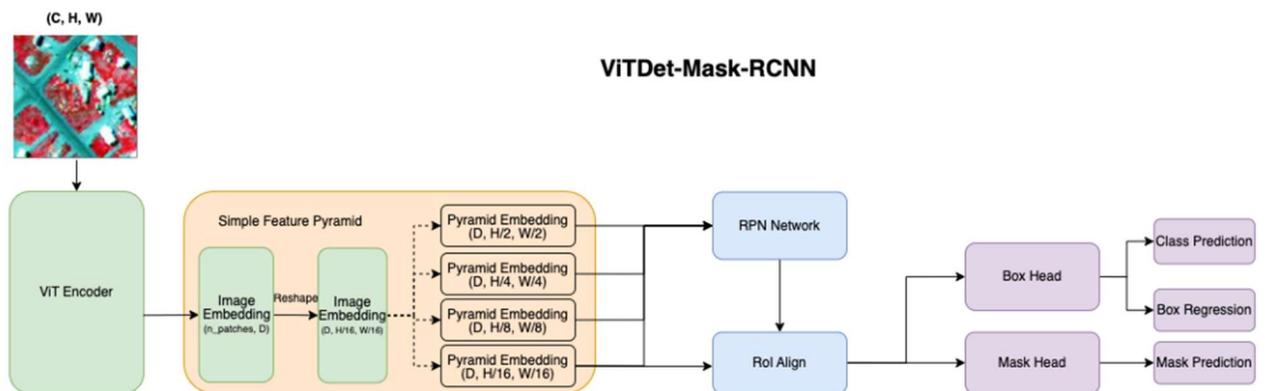

**Figure 3.** The top half of the figure shows a high-level representation of the Mask-RCNN framework. The bottom half of the figure showcases how the framework can be adapted to work with the feature maps produced by a SFPN instead of relying on a hierarchical backbone.

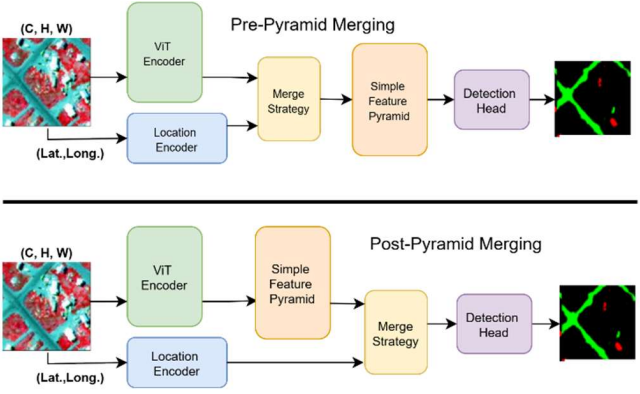

**Figure 4.** Pre and Post (SFP) Location Encoder placement with multiple merge strategies.

In addition to tiling the 1D location embedding to produce a 2D map that can be merged with the image embeddings as described above, we also introduce the idea of projecting the 1D embedding into a 2D raster that can then be added across channel dimensions to the image embeddings or appended as an extra channel (projection add on or projection concatenation respectively). For $F$ and $L$, we apply these projection operations to produce $C$ as given in equations (7 - 8) where $W_l L$ indicates the projection of $L$ with learnable projection weights $W_l$ in addition to the previously defined terminology:

$$C = F + L_p \text{ where } L_p = W_l L \quad (7)$$

$$C = Concat(F, L_p) \text{ where } L_p = W_l L \quad (8)$$

Lastly, we also introduce a form of cross attention mechanism that allows the model to dynamically and contextually align information from feature embeddings with location embeddings. By leveraging attention scores, the fused representation captures nuanced interactions between the image features and spatial locations. This could lead to embeddings that are both rich in spatial semantics and aligned with visual content. For $F$ and $L$, we apply cross attention to produce $C$ as given in equation (9) where $W_q F$ is the query projection of the input features $F$, $W_k L$, $W_v L$ are the key and value projections of the input L using learnable $W_k$ and $W_v$, $Softmax\left(\frac{QK^T}{\sqrt{d}}\right)$ is the attention weight matrix computed using scaled dot-product attention, where $d$ is the dimensionality of the key vectors, in addition to the previously defined terminology:

$$C = AV \text{ where } Q = W_q F, K = W_k L, V = W_v L, \text{ and } A = Softmax\left(\frac{QK^T}{\sqrt{d}}\right) \quad (9)$$

## IV. EXPERIMENTS

### A. Training Datasets

Here, we introduce the datasets that are used to empirically design the model and demonstrate the capabilities and improvements of the proposed model for feature detection using ViTs with location embeddings. We chose three different data sets for three different target objects, which were previously created for prior studies using CNN-based models on the Arctic region. This will enable us to demonstrate the model's ability to adapt to different Arctic feature detection downstream tasks while providing improved accuracy compared to the prior CNN-based models. The training data sets selected from prior studies are for Regressive Thaw Slumps (RTS) [58], Infrastructure in the Arctic region [59], and Ice-wedge Polygons (IWP) [60]. We ensure that the validation data sets are used for the empirical model design process through architectural ablations, and the test set is kept independent to report the final metrics for the respective datasets.

Each dataset was created within the *ArcGIS* software for prior studies through the same on-screen digitization process with Maxar satellite imagery, which has a spatial resolution of < 1 m. This imagery was pre-processed (pansharpened, orthorectified) and provided by the Polar Geospatial Center at the University of Minnesota. Analysts were tasked with visually identifying and digitizing the outline of a target object's footprint (i.e., RTS, IWP, or Infrastructure) in each Maxar satellite image. See [59], [60] or [58], for examples of what specific criteria were followed to digitize target objects. Imagery acquired during the summer with minimal cloud cover was selected for training dataset creation and a false color composite (near infrared, red, and green image bands) was used during digitization.

The RTS training dataset is constrained to the Canadian Arctic Archipelago, the infrastructure and IWP training datasets are composed of satellite imagery acquired across an expansive circumpolar geographic domain that accounts for variability in the built environment (for infrastructure) and tundra types, or lack thereof (for both infrastructure and IWPs). These different environments present a range of features unique to geographic locations that a model may need to learn. For example, two different tundra types with different vegetation will exhibit different spectral reflectance; urban centers contain much more development than rural villages, which could result in more spectral variability in imagery due to wider range of construction materials. Accounting for this variability is imperative in developing a robust segmentation model that works on a regional scale. To get an idea of the geographical diversity and dispersion of the training data we show the geographic locations of the training data sets used for the experimentation in Figure 5.

*Retrogressive thaw slumps*

To train and evaluate semantic segmentation models for RTS detection, training data was collected from satellite imagery of two study sites (Figure 5): (1) Banks Island (~3700 km2), Northwest Territories, and (2) Axel Heiberg Island and Ellesmere Island, referred to as the Eureka Sound Lowlands (ESL), (~4000 km2), Nunavut, Canada. RTS objects were manually annotated from Worldview-02 (WV-02) imagery that were acquired during the summertime between the years 2010 and 2015 (Banks Island) and 2011 and 2020 (Ellesmere). Banks Island is the westernmost island, while ESL consists of the two northernmost islands in the Canadian Arctic Archipelago. 12 satellite image scenes were used for Banks Island, while 14 were used for Ellesmere Island, all having spatial resolutions of 0.5 m.

In total, 950 RTS objects were digitized (n=475 from each site) (see Table 1 for Summary statistics). Box plots of RTS

geometric properties, such as area, length (major axis), width (minor axis), and length-to-width ratio describe the basic size and shape characteristics of the hand-annotated RTS samples from Banks Island and ESL (Figure A1(b) in the appendix). RTS training samples from Banks Island and the ESL reported median areas of 7.03 ha and 4.67 ha, respectively. Furthermore, RTS samples from Banks Island showed higher variability in area, length, and width compared to the samples from ESL [58]. A total of 2,132 image tiles were distributed as follows: 1705 for training, 214 for testing, and 213 for validation, each with a tile size of 1024×1024 pixels.

TABLE I
SUMMARY STATISTICS OF THE DIGITIZED RTS TARGET OBJECTS.

| Study area | No. of Objects |
|---|---|
| Banks Island | 475 |
| Ellesmere Island | 475 |

*Infrastructure*

The infrastructure dataset was strategically created to account for regional variability in the natural and built environment across the Arctic. 25 Maxar satellite images (WorldView-02 and -03, and QuickBird-02) over 18 different sites across Arctic Alaska, Canada, and Russia (Figure 5) were used [16]. Each of these sites represents a particular built environment setting (either a rural settlement, medium-density settlement, urban settlement, or industrial site) and climate setting (either tundra or boreal climate).

This diversity ensures that a deep learning model learns the variability inherent to infrastructure built across different settings. For example, infrastructure detection in boreal climates is challenged by tree occlusions that block buildings or roads; this is not seen in tundra landscapes. Urban infrastructure detection is challenged by overlapping buildings due to dense spatial distributions, which also often take on irregular shapes. Urban areas are also characterized by heterogeneous development that consists of various surface materials (e.g., concrete, brick, asphalt, metal, plastic, glass, shingles, and soil). This spectral variability is problematic for detection. Figure A1(a) in the appendix summarizes the distribution of various geometric properties of the dataset. A total of 5,374 image tiles were distributed as follows: 4,362 for training, 506 for testing, and 506 for validation, each with a tile size of 256×256 pixels.

TABLE II
NUMBER OF LABELED INSTANCES OF EACH INFRASTRUCTURE CLASS IN THE DATASET.

| Country | Buildings | Roads | Storage tanks |
|---|---|---|---|
| U.S. (Alaska) | 2,518 | 552 | 73 |
| Canada | 3,692 | 508 | 85 |
| Russia | 885 | 109 | 59 |
| Total | 7,095 | 1,169 | 217 |

*Ice-wedge polygons*

The IWP training data set was sampled from a few representative Maxar satellite image scenes captured between 2012 and 2015 across various locations in the circumpolar Arctic Figure 5 [60]. Table III shows the distribution of the training samples across Arctic Alaska, Canada, and Russia that includes almost 33,000 IWPs on 855 image tiles, split as 609 for train, and 123 each for test and validation with tile sizes ranging from 199x199 pixels to 504x504 pixels. The same experimental design rigor described earlier for the creation of the RTS and Infrastructure datasets was used to create the IWP dataset. In each image tile, IWP outlines were manually digitized and labeled as "low" for low-centered or "other" based on their morphometry. Figure A1(c) in the appendix summarizes the distribution of various geometric properties of the dataset.

TABLE III
NUMBER OF LABELED INSTANCES OF EACH IWP CLASS IN THE ENTIRE DATASET.

| Country | Low | Other | Total |
|---|---|---|---|
| U.S. (Alaska) | 9,540 | 10,569 | 20,109 |
| Canada | 1,968 | 3,228 | 5,196 |
| Russia | 1,804 | 5,982 | 7,786 |
| Total | 13,312 | 19,779 | 33,091 |

As stated earlier the three feature detection datasets used for this study were selected from prior studies and display different characteristics with respect to the dataset size and the geographic distribution enabling us to infer interesting design considerations and generalization about the model architecture proposed in this work.

*B. Experimental setup*

We leverage the Detectron2 [61] open-source framework developed by Facebook AI Research (FAIR) for implementing state-of-the-art CV models to empirically build and refine vision models for image segmentation. We were able to incorporate domain-specific knowledge, such as geospatial context via location embeddings, and TransUNets with SFPN for the detection head using the framework. Detectron2 allows for easy customization of models and training pipelines allowing us to seamlessly modify components like the backbone, detection heads, and data loaders to suit specific tasks. Detectron2's modular design makes it straightforward to integrate new features like location embeddings into the pipeline. It also supports efficient distributed data parallel (DDP) multi-GPU training, making it suitable for large-scale vision tasks, such as processing RS data with high spatial resolution.

The computing infrastructure used for this work includes U.S. NSF-funded Texas Advanced Computing Center (TACC) Peta-scale HPC Frontera [62] with 4 NVIDIA Quadro RTX 5000 GPUs with 16GB GDDR6 VRAM per node, HPC Lonestar [62] with 3 A100 GPUs with 40GB HBM2 per node [63], and Google Cloud Platform (GCP) virtual machines with 8 NVIDIA L2 with 24 GB GDDR6 per node and A100 GPUs with 40GB HBM2 per node [64].

During experiments, models were trained for 75 epochs to ensure convergence. The number of iterations per epoch was determined by dividing the total training dataset size by the effective batch size. The effective batch size was calculated as the product of the per-GPU batch size and the number of GPUs used [65]. The per-GPU batch size for each task was set to the maximum possible for the given compute architecture to optimize memory and computational efficiency. The resulting number of iterations for each experiment is reported in the experimental results as shown in Table IV. Note that the input image tile size for the IWP training data set varied from 199 to 504 and the batch size was selected to suite the largest input

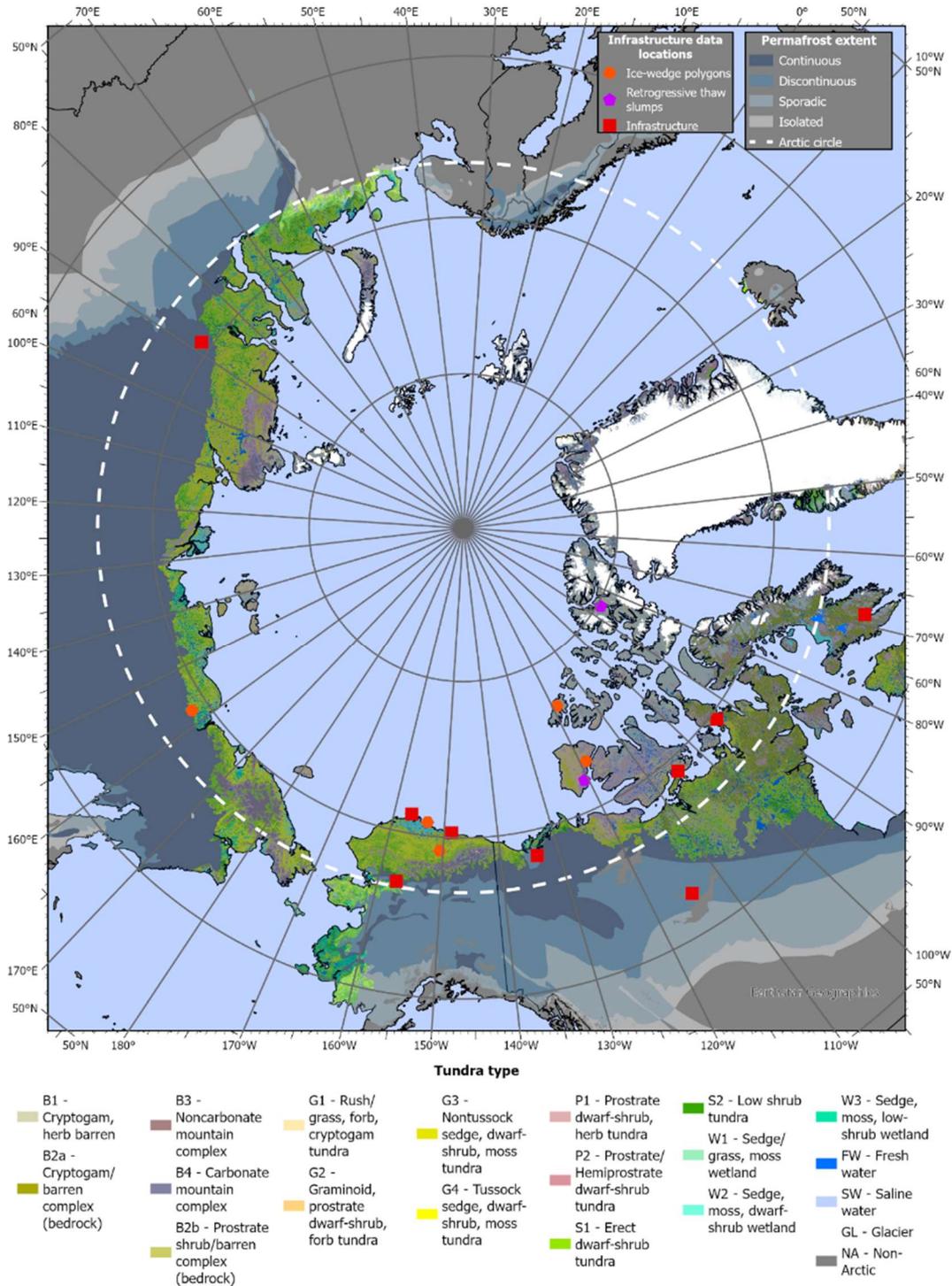

**Figure 5.** Sampled locations from which target objects for each dataset were labeled from Maxar satellite imagery. Distribution of tundra types based on the Circumpolar Arctic Vegetation Map [66] and permafrost extent based on the Circum-Arctic map of permafrost and ground-ice conditions [67].

image tile. We also leveraged large scale jittering with the default settings as described in [68] to help with data augmentation. This was especially relevant to prevent overfitting given the relatively small dataset sizes used in this work. Where feasible, multiple validation trials were performed to account for variability, and mean accuracy values were used for a reliable assessment of model performance comparison

TABLE IV
NUMBER OF ITERATIONS USED FOR EACH COMPUTE ARCHITECTURE

| | | | Infra | RTS | IWP |
|---|---|---|---|---|---|
| Number of samples → | | | 4306 | 1706 | 606 |
| Input image tile size in pixels → | | | 256 | 1024 | *512 |
| model | gpu's | mem(GB) | | Iterations | |
| **RTX 5000** | 4 | 16 | 20500 | 8000 | 2900 |
| **A100** | 1 | 40 | 2600 | 4000 | 5700 |
| **A100** | 3 | 40 | 900 | 1350 | 1900 |
| **L2** | 8 | 24 | 1703 | 16000 | 710 |

## V. RESULTS AND DISCUSSION

### A. Architectural Ablations

We present the evaluation results of multiple ablation experiments that support the proposed empirically designed architecture for feature detection using ViTs with location embeddings. Table V shows the overall experimental dimensions and the different options considered for the ablations.

TABLE V
ABLATION STUDY DIMENSIONS

| ViT Encoder | Geo. Encoder | Placement | Merge Strategy (Eqn) |
|---|---|---|---|
| Base | L10/L40 | Post Pyramid | Add (2) |
| Base | L10/L40 | Post Pyramid | Norm Add (4) |
| Base | L10/L40 | Pre/Post Pyramid | Concat (3) |
| Base | L10/L40 | Pre/Post Pyramid | Norm Concat (5) |
| Base | L10/L40 | Pre/Post Pyramid | Concat Norm (6) |
| Base | L10/L40 | Pre/Post Pyramid | Proj Add (7) |
| Base | L10/L40 | Pre/Post Pyramid | Proj Concat (8) |
| Base | L10/L40 | Pre/Post Pyramid | Cross Atten. (9) |
| Base/Large/Huge | - | - | - |

*Backbone Size.*

Larger backbones typically offer richer feature representations due to their increased depth and parameter count, potentially improving accuracy on complex tasks. However, this comes at the cost of higher computational and memory requirements, which may not be practical for all deployment scenarios. Conversely, smaller backbones are more lightweight and efficient but may sacrifice performance, especially on tasks requiring fine-grained detail or high contextual understanding. By systematically evaluating models with different backbone sizes, we can empirically determine the optimal balance for the specific task of arctic feature detection, guiding the selection of architectures that maximize performance. We test our proposed architecture with three pre-trained ViT backbones, base (86 million parameters), large (307 million parameters), huge (632 million parameters) available on Detectron2 [61] for the three different training datasets and report the validation metrics. All three size variations were pre-trained on the ImageNet-1K dataset using a MAE approach [56]. The backbone weights were not frozen and were allowed to adapt to the new training samples considering the difference between ImageNet data and the training data used for this work. A comparison of results obtained for all three datasets without location embeddings is given on Table VI, Table VII, and Table VIII, and with detailed results, including error margins, given in the appendix. When reporting results we use standard definitions for Semantic Segmentation [69] and Instance Segmentation [70] metrics.

TABLE VI
VALIDATION METRICS FOR BACKBONE SIZE FOR RTS

| ViT Size | Pix. Acc. | Prec. | Rec. | F1 | mIoU |
|---|---|---|---|---|---|
| Base (86m) | 0.9406 | 0.8217 | 0.7973 | 0.8086 | 0.7091 |
| Large (307m) | 0.9222 | 0.8278 | 0.7463 | 0.7781 | 0.6728 |
| Huge (632m) | 0.9401 | 0.8320 | 0.7932 | **0.8110** | **0.7117** |

TABLE VII
VALIDATION METRICS FOR BACKBONE SIZE FOR INFRASTRUCTURE

| ViT Size | Pix. Acc. | Prec. | Rec. | F1 | mIoU |
|---|---|---|---|---|---|
| Base (86m) | 0.9507 | 0.7888 | 0.8027 | **0.7939** | **0.6751** |
| Large (307m) | 0.9463 | 0.8204 | 0.7491 | 0.7674 | 0.6437 |
| Huge (632m) | 0.9476 | 0.8358 | 0.7724 | **0.7947** | **0.6755** |

TABLE VIII
VALIDATION METRICS FOR BACKBONE SIZE FOR IWP

| ViT Size | mAP | $mAP_{50}$ | $mAP_{75}$ | $AP_s$ | $AP_m$ | $AP_l$ |
|---|---|---|---|---|---|---|
| Base (86m) | 0.2506 | 0.4820 | 0.2474 | 0.1944 | 0.4484 | 0.5754 |
| Large (307m) | **0.2721** | **0.5077** | 0.2764 | 0.2158 | 0.4724 | 0.6161 |
| Huge (632m) | 0.2688 | 0.5020 | 0.2766 | 0.2103 | 0.4651 | 0.6005 |

As anticipated for the Infrastructure data and the RTS data, the huge backbone variant of the ViT encoder surpasses the performance of the other two encoder configurations. However, it is important to note that the improvement in the F1 score is not substantial. In contrast for the IWPs both the large and huge variants of the ViT encoders significantly surpass the performance of the base variant for $mAP_{50}$. The findings from encoder variations provide a compelling illustration of the principle that model complexity does not invariably correlate with superior performance. Based on the observations we can safely select the base backbone for the RTS and Infrastructure data sets and the large backbone for the IWP dataset.

*Location Embeddings*

As discussed in section III.B we explore the architectural variants based on the placement, merge strategy and the encoder granularity of the location embeddings produced by the SatCLIP model to empirically identify the best design configuration for the respective downstream task. All the detailed experimental results are presented in Appendix A and in this section, we highlight in Table IX, Table X, Table XI significant results that strongly influence the design decisions for the final model

For the infrastructure dataset, we observed that the L10 post-pyramid projection concatenation model yielded the highest overall performance, achieving a mean Intersection over Union (mIoU) of 0.8080 and an F1 score of 0.8901. One interesting insight is that although there doesn't seem to be an evident clear winner between using the L10 and L40 variations of the location embedding model, most of the L10 models seem to be performing better than the L40. One likely reason for this is that the sites selected to compile the infrastructure dataset are spatially dispersed meaning that even the coarser L10 embedding model can provide meaningful enough location information. The opposite of this can be seen below for the RTS dataset.

The analysis of location embedding variations within the RTS dataset, as previously alluded to, underscores the efficacy of the more granular L40 embeddings when applied to datasets characterized by spatial proximity and homogeneity. Across all architectural variations examined, the models incorporating L40 embeddings consistently outperformed their L10 counterparts. Notably, the pre-pyramid L40 concatenation and the projection concatenation exhibit the highest performance within this dataset with a few other L40 variants also showing significant performances with respect to the mIoU and the F1 scores.

The results above highlight a noteworthy influence of location embedding granularity. Initially, the IWP dataset appears to share a characteristic with the infrastructure dataset: a distribution of distinct sites across the Arctic region. Based on this apparent similarity, one might anticipate a comparable outcome to the infrastructure dataset analysis, where the choice of L10 or L40 location embeddings did not significantly affect model performance. However, sites within the IWP dataset can be more accurately characterized as clustered distributions dispersed across the Arctic. This observation offers a plausible explanation for the superior performance observed with the L40 models. The finer granularity of the L40 embeddings enables the model to not only differentiate between sites situated in geographically distant regions but also to discern subtle spatial variations between relatively proximate sites. This enhanced capacity for local discrimination appears to contribute to the observed performance advantage. Resulting in a $mAP_{50}$ of 0.4917 for the L40 pre-pyramid cross attention.

Table XII shows the selected model configurations for the respective downstream tasks, and the selected model configuration that will be used to compare the models using the test datasets presented later.

*Impact of Location Embeddings*

Figure 6 shows the comparison of model accuracy of the best model configuration (Table XII) with location embeddings, compared with the model without location embeddings and the existing CNN model [16],[58],[60] for the respective dataset. As can be seen there is an increase in the accuracy of the models with location embeddings compared to the other models. To remove any potential variability introduced due to using different ViT encoder sizes we use the base ViT model across all datasets for this comparison.

TABLE IX
TOP VALIDATION METRICS FOR FOR INFRASTRUCTURE DATA

| Configuration (placement/granularity/merge strategy) | Pixelwise accuracy | Precision | Recall | F1 score | mIoU |
|---|---|---|---|---|---|
| Post L10 Proj Concat | 0.9659 | 0.8740 | 0.9074 | 0.8901 | 0.8080 |
| Pre L10 Cross Attention | 0.9656 | 0.8848 | 0.8881 | 0.8863 | 0.8019 |
| Pre L10 Norm Concat | 0.9655 | 0.8808 | 0.8856 | 0.8831 | 0.7968 |
| Post L40 Concat | 0.9649 | 0.8705 | 0.8960 | 0.8828 | 0.7966 |
| Post L40 Cross Attention | 0.9655 | 0.8719 | 0.8931 | 0.8822 | 0.7955 |
| Pre L10 Proj Concat | 0.9652 | 0.8761 | 0.8882 | 0.8821 | 0.7955 |

TABLE X
TOP VALIDATION METRICS FOR FOR RTS DATA

| Configuration (placement/granularity/merge strategy) | Pixelwise accuracy | Precision | Recall | F1 score | mIoU |
|---|---|---|---|---|---|
| Pre L40 Concat | 0.9554 | 0.8488 | 0.8507 | 0.8498 | 0.7599 |
| Pre L40 Proj Concat | 0.9568 | 0.8345 | 0.8642 | 0.8486 | 0.7587 |
| Pre L40 Concat Norm | 0.9565 | 0.8343 | 0.8627 | 0.8478 | 0.7577 |
| Post L40 Proj Add | 0.9553 | 0.8417 | 0.8530 | 0.8472 | 0.7567 |
| Pre L40 Norm Concat | 0.9493 | 0.8654 | 0.8210 | 0.8410 | 0.7481 |
| Pre L40 Cross Attention | 0.9529 | 0.8351 | 0.8447 | 0.8397 | 0.7473 |

TABLE XI
TOP VALIDATION METRICS FOR FOR IWP DATA

| Configuration (placement/granularity/merge strategy) | mAP | $mAP_{50}$ | $mAP_{75}$ | $AP_s$ | $AP_m$ | $AP_l$ |
|---|---|---|---|---|---|---|
| Pre L40 Cross Atten | 0.2548 | 0.4917 | 0.2533 | 0.2001 | 0.4446 | 0.575 |
| Post L40 Norm Concat | 0.2545 | 0.4907 | 0.2519 | 0.2004 | 0.4445 | 0.5663 |
| Post L10 Proj Concat | 0.2531 | 0.4854 | 0.2530 | 0.1978 | 0.4455 | 0.5989 |
| Pre L10 Proj Add | 0.2532 | 0.4851 | 0.2519 | 0.1986 | 0.4428 | 0.5664 |
| Pre L10 Concat | 0.2534 | 0.4849 | 0.2547 | 0.1987 | 0.4465 | 0.5801 |

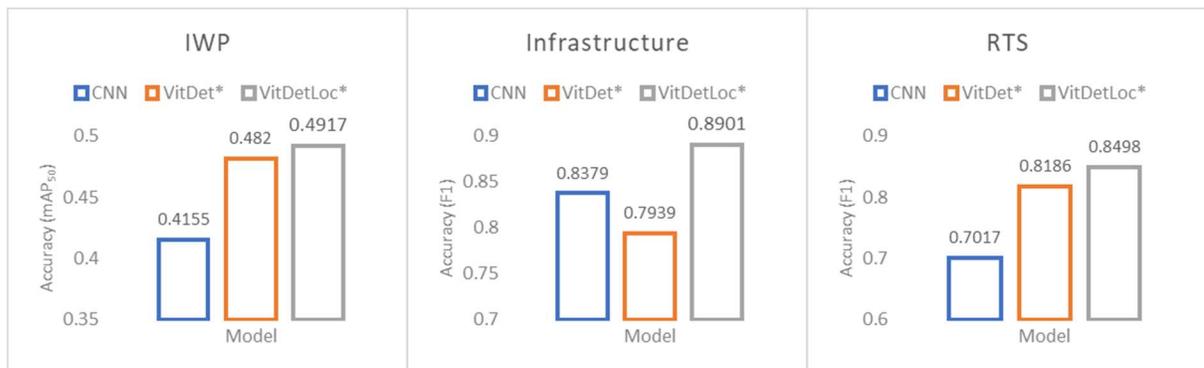

**Figure 6.** Comparison of the impact with location embeddings and without location embeddings shows a significant improvement of model accuracy for the models with location embeddings for all three mapped features. For IWP model* indicates ViTDet-Mask-RCNN and for Infrastructure and RTS it is ViTDet-UNet

TABLE XII
SELECTED MODEL CONFIGURATION FOR THE DOWNSTREAM TASK BASED ON VALIDATION RESULTS

| Downstream Task | ViT | Encoder Placement | Location Encoder | Merge Strategy |
|---|---|---|---|---|
| Infrastructure | Base | Post | L10 | Proj Concat |
| RTS | Base | Post | L40 | Concat |
| IWP | Large | Pre | L40 | Cross Attention |

The t-Distributed Stochastic Neighbor Embedding (t-SNE) is a nonlinear dimensionality reduction technique used to visualize high-dimensional data by mapping it to a lower-dimensional space while preserving local similarities [36]. We apply t-SNE to project the pre-trained location embeddings generated by SatCLIP into a 2D space, enabling qualitative assessment of their spatial structure and separability. Figure 7 includes the t-SNE visualization of the SatCLIP embeddings using the L40 encoder with perplexity at 50 for 5000 iterations for a subset of the infrastructure training data with the geographic locations marked on the accompanying map. Although cluster distance and sizes don't mean much in a t-SNE visualization, the fact that the different sites create independent clusters indicates that the location embedding model is providing a unique signal for each site. If the embeddings for the various sites were all part of a big cluster, then the signal coming from the location embedding model would not be useful in differentiating the unique characteristics of each location. As it was seen in the previous comparison of the models with and without location embeddings the location embeddings can be attributed for enhancing the downstream detection by fusing location dependent information into the model.

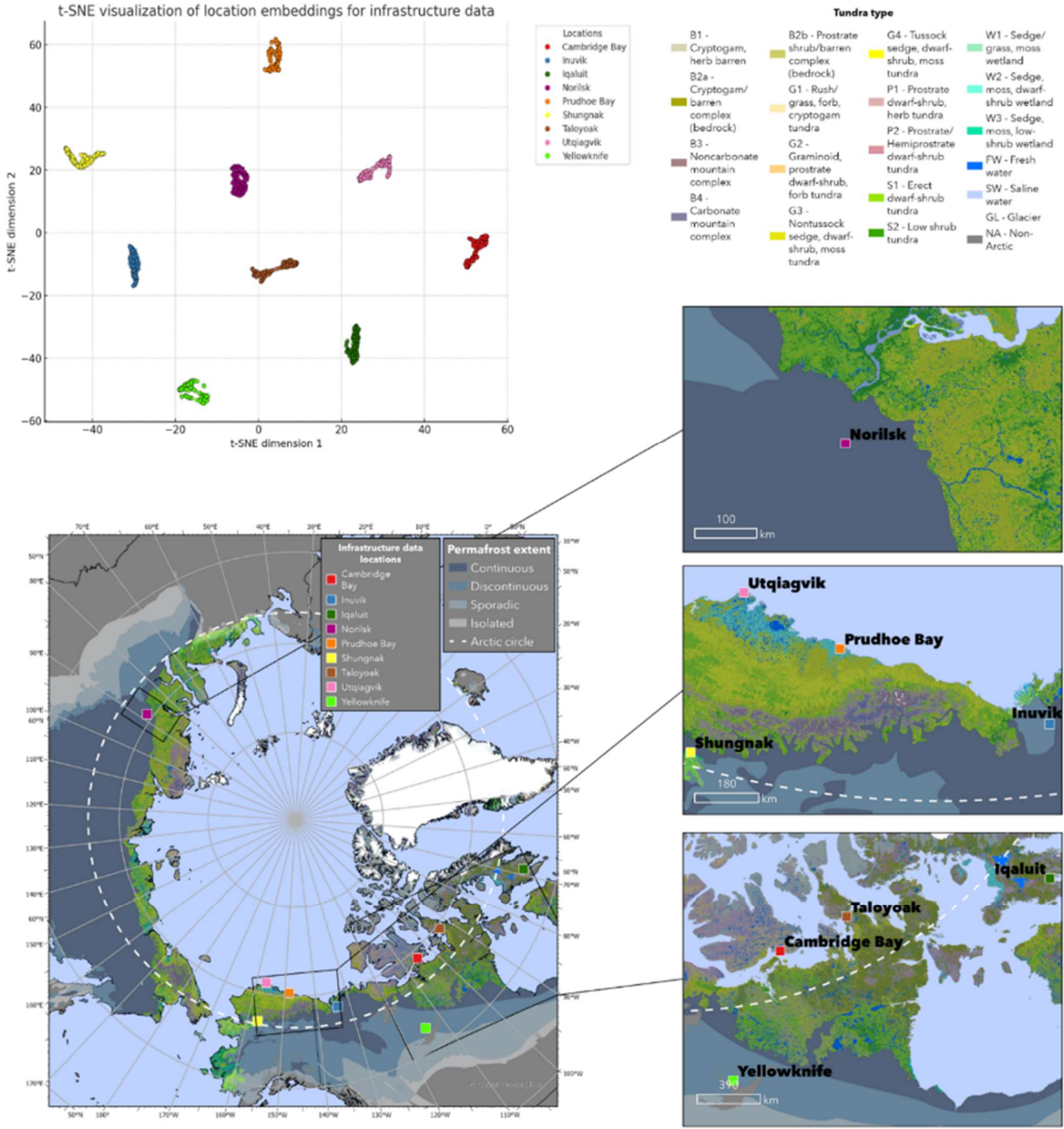

**Figure 7.** t-distributed stochastic neighbor embedding (t-SNE) visualization of the SatCLIP embeddings using the L40 encoder for a subset of the infrastructure training data with independent clusters for each site.

### A. Performance Comparison with Best Configuration

Table XIII presents the comparison of the test results for the proposed best model configurations against the existing CNN based models for RTS [58], Infrastructure [16], and IWP [60]. In addition, we also compare the results with another transformer-based model, Mask2Former [47] that has been introduced as a single model that is capable of handling semantic, instance and panoptic segmentation. The configuration of pre-trained Mask2Former model used for comparison employs a Swin Transformer backbone built with a 384 × 384 input resolution, a batch size of 16, and is trained for 50 epochs. The pre-trained Mask2Former model was trained on our downstream tasks multiple times with a few basic parameter variations and the best model is used for comparison.

For each test dataset our proposed, empirically designed, models with location embeddings show significantly better results for both IWPs and RTS when compared to the other candidate models except for the infrastructure dataset although for the validation dataset the proposed model outperforms for all 3 categories as shown in Figure 6. Also there is a difference in the performance values for the test and the validation datasets which can be attributed to the very small sizes of the validation and test sets. The best model configuration for each dataset was selected based on the results of the ablation studies discussed in section V.A Following standard practices, we make all model architecture decisions based on observed results from the validation dataset and hold out the test dataset exclusively for reporting the results shown below. This helps simulate the expected results when using the respective models on downstream feature detection tasks on a large volume of satellite image tiles from the pan-Arctic region.

TABLE XIII
TEST DATA PERFORMANCE COMPARISON

| Dataset | IWP | Infrastructure | RTS |
|---|---|---|---|
| Metric | mAP$_{50}$ | F1 | F1 |
| Model w/Loc. Emb. | **0.5722** | 0.8604 | **0.9239** |
| CNN based | 0.4968 | **0.8887** | 0.8455 |
| Mask2Former | 0.5034 | 0.5615 | 0.6336 |

We used the default settings provided for the Detectron2 ViTDet parameters, as our primary objective was to evaluate the impact of Vision Transformers and location embeddings on Arctic feature detection. Although this configuration was sufficient for our exploratory analysis, additional performance improvements are likely achievable through hyperparameter tuning, which we strongly recommend prior to deploying these models in large-scale inference pipelines. Notably, since the CNN-based baselines used for comparison were fine-tuned, the performance achieved by our proposed model, despite using untuned default settings, can be considered even more favorable.

### B. Example Detections for the Datasets

We present a compilation of examples from the respective datasets with the original image tile in false color composite, and the annotated ground truth for three datasets used in this work. We also discuss the performance of the model on the respective image tiles and the possible contribution from the location embeddings. Figure 8 for infrastructure dataset illustrates how model with location embeddings can improve the detection for image tiles from the dataset with very low training representation based on the location distribution of the annotated training samples.

Figure 9 illustrates how RTS identification is improved by the proposed model with location embeddings by reducing False Positives (FP) compared to the model without location embeddings. Figure 10 illustrates how IWP classification is improved by the proposed model with location embeddings compared to the base model without location embeddings.

Our experimental results across three challenging Arctic feature detection tasks—ice-wedge polygons, retrogressive thaw slumps, and human infrastructure—demonstrate that transformer-based models not only match but often surpass CNN-based baselines. The ViT-based models proved effective in capturing long-range dependencies and global spatial context, which is critical for dense prediction tasks in high-resolution satellite imagery.

To address the issue of spectral variability in semantically similar features across geographically diverse regions with limited labeled training data, a common challenge in Arctic remote sensing, we introduced geospatial location embeddings. These embeddings provided models with contextual awareness of geographic position, improving their ability to disambiguate visually similar features based on location. Even with the limited training data we were able to demonstrate the ability of the model with location embedings to perform better than models without location embedgings for under represented areas. Comparative evaluations of different integration strategies revealed that architectural choices, such as the fusion point, merging method, and spatial resolution of the embeddings, influence detection accuracy.

The results also confirm that replacing the CNN backbone in TransUNet with a transformer and multiscale feature pyramid (SFPN) preserves hierarchical feature extraction.

These results confirm the value of combining ViTs with semantically informed location embeddings, offering a robust approach for scalable and accurate Arctic remote sensing.

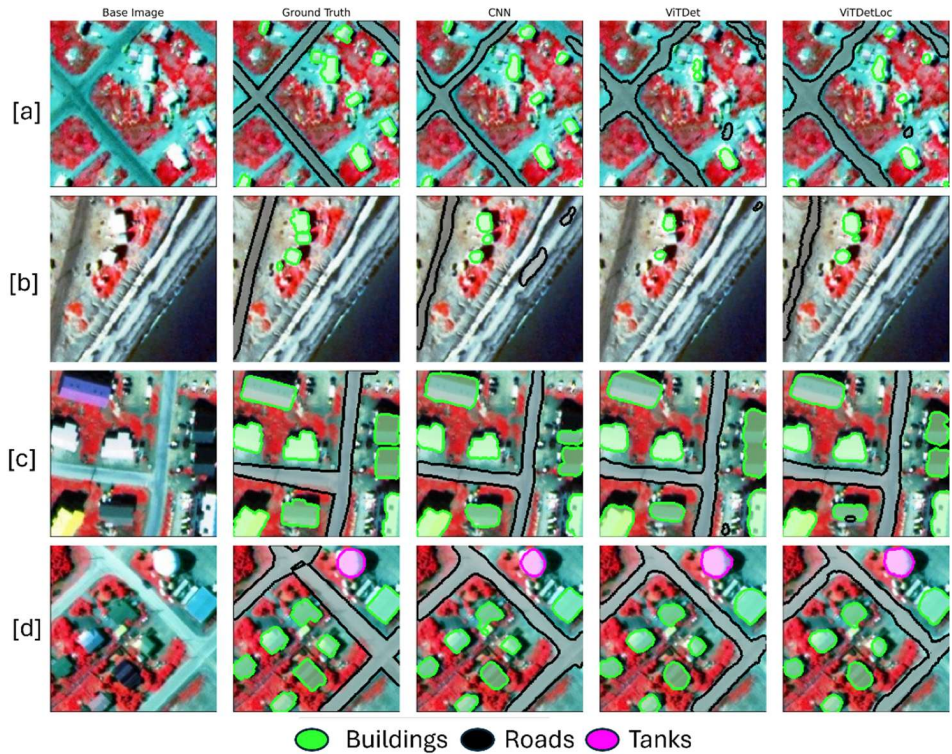

**Figure 8.** Segmentation model results for four example tiles (a-d), showing base image (Satellite Images © Maxar), annotated ground truth, and outputs for the different models. Green outlines represent buildings, black represent roads, and magenta represents storage tanks. [a] Tile from Wainwright, which makes up about 1% of the training data. The model with location embedding (ViTDetLoc*) shows a noticeable improvement in segmentation compared to the other models. [b] Another tile from the Wainwright, where the ViTDetLoc* model correctly identifies the road, highlighting its better generalization in underrepresented areas compared to the ViTDet model. [c] Both the CNN and ViTDetLoc* segment the ground truth reasonably well. In comparison ViTDet* struggles to segment individual buildings when compared to the VitDetLoc* results. [d] The ViTDetLoc* model can more accurately capture the shape of segmented objects. Both models perform reasonably well overall, but the location embedding provides a closer object outline to the ground truth. Note: Model* indicates ViTDet-UNet.

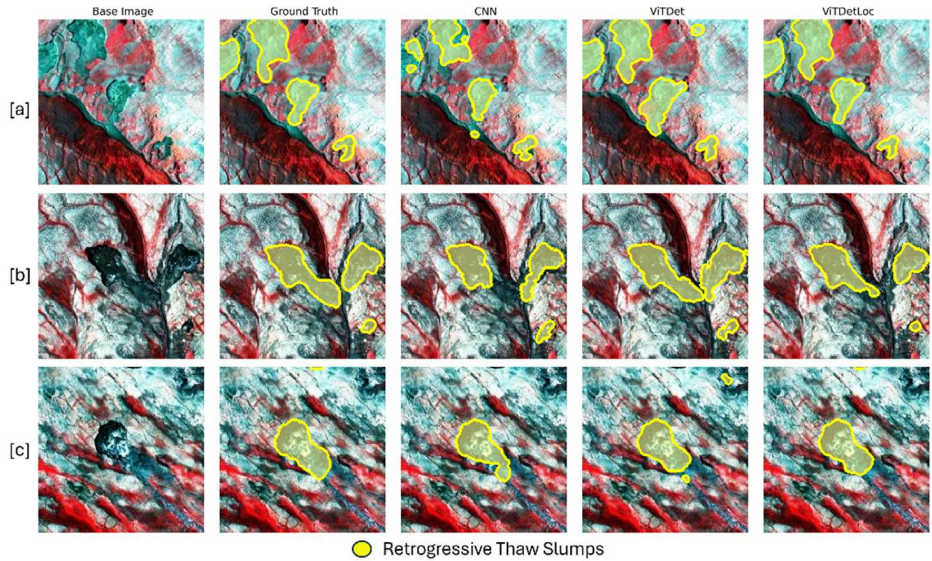

**Figure 9.** Segmentation model results for three example tiles (a-c), showing base image (Satellite Images © Maxar), annotated ground truth, and outputs from the CNN, VitDet, and VitDetLoc models. Yellow outlines represent the annotated retrogressive thaw slumps. [a] Both VitDet and VitDetLoc models segment the thaw slumps reasonably well, although the VitDet model includes a false positive not present in the ground truth. The ViT Models show greater performance than the CNN in overall segmentation [b] Segmentation is generally accurate across both VitDet and VitDetLoc models, however the VitDetLoc model shows more conservative segmentation results than VitDet. The ViT Models show greater performance than the CNN in overall segmentation. [c] Both VitDet and VitDetLoc models align closely with the ground truth, but the VitDet model includes a false positive while missing a true positive, that the VitDetLoc model avoids. Overall, both the VitDet and VitDetLoc outperform the CNN. Note: Model* indicates ViTDet-UNet.

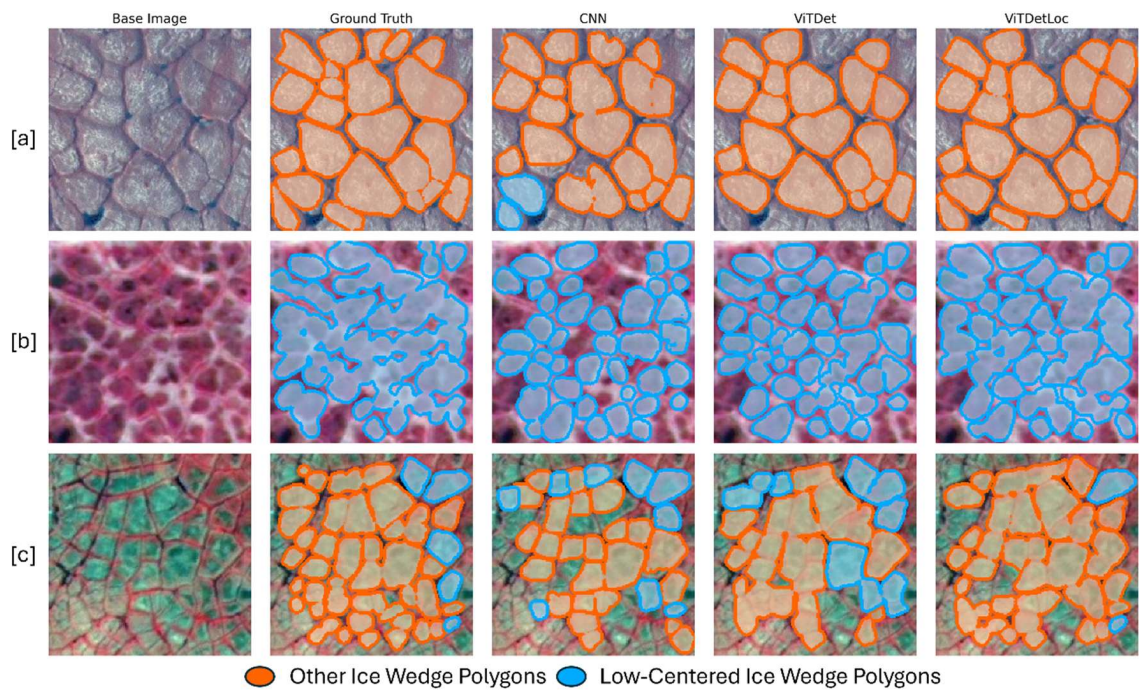

**Figure 10.** Segmentation model results for three example tiles (a-c), showing base image (Satellite Images © Maxar), annotated ground truth, and outputs from the CNN, VitDet, and VitDetLoc models. Blue outlines represent the annotated low centered ice-wedge polygons and the orange outlines represent other ice-wedge polygons. [a] Both ViTDet and ViTDetLoc perform well, though the VitDet model without location embedding misses a few true positives. Both ViT models perform better than the CNN, as the models predict more true positives. Additionally, the CNN has two low centered polygon classifications that are incorrect which the ViT's correctly identify. [b] Both ViTDet and ViTDetLoc models segment the ground truth accurately overall, but the VitDet model fails to detect a few true positives. However similar to (a) the CNN misses some true positives which the ViT's capture [c] Both ViTDet and ViTDetLoc models capture most of the segmentations correctly, but the ViTDet model shows a few misclassifications. The CNN misses many polygons and has several misclassifications. Note: Model* indicates ViTDet-Mask-RCNN..

## VI. CONCLUSION

In this work, we explored and enhanced the capabilities of remote sensing in the Arctic permafrost region by leveraging the latest advancements in CV, particularly ViTs, in conjunction with semantically rich location information. By integrating these modern deep learning techniques, we improved the detection accuracy and characterization of pan-Arctic features, which are crucial for environmental monitoring, resource management, and accurate risk assessment. Our findings demonstrate that pre-trained ViTs can be successfully adapted for pan-Arctic feature detection and segmentation, showcasing their effectiveness in extracting meaningful spatial patterns from high resolution satellite imagery. In this work we also show how we can leverage the multiscale feature map produced by a SFPN in place of the CNN in the TransUNet architecture.

Furthermore, we highlight the significant role of location embeddings in improving detection performance. Specifically, incorporating SatCLIP, a pre-trained location embedding model, enhances the model's ability to contextualize geospatial information, leading to more accurate and robust predictions.

Additionally, our experiments reveal that the effectiveness of location embeddings is influenced by key architectural choices, including encoder placement, merging strategy, and embedding granularity. These factors directly impact the downstream detection task, emphasizing the need for carefully designed fusion mechanisms when integrating geospatial information with visual features.

Overall, our study underscores the potential of combining ViTs with geospatial embeddings to advance remote sensing applications in Arctic environments. By refining the integration of location-aware representations, we pave the way for more precise and scalable solutions in large-scale geospatial monitoring, ultimately contributing to better decision-making in Arctic scientific research and policy.

*Future Work*

Building on our findings, several key directions can be explored to further enhance the integration of ViTs with geospatial embeddings for remote sensing applications in the Arctic and beyond. Two ideas we plan to pursue are; (1) Although in this work we use Cross Attention and Learned Projections for merging Location Embeddings with acceptable results, there is more in-depth analysis that can be done to enhance these two approaches. (2) With the availability and access to a large volume of commercial high resolution satellite images and the lack of pre-trained ViTs specifically for the pan-Arctic regions, it is prudent to build our own ViT using MAE, including extra bands, and a suitable input tile size for downstream arctic detection tasks, that leverages larger images than the 224 x 224 used by the pre-trained ViTs in this work.

## APPENDIX

Figure A1 depicts the feature characterization of the respective training datasets. Feature compactness is computed as given in equation (10) for the figures A1(a),(b), and (c)

$$Compactness = \frac{Perimeter^2}{4\pi \times area} \quad (10)$$

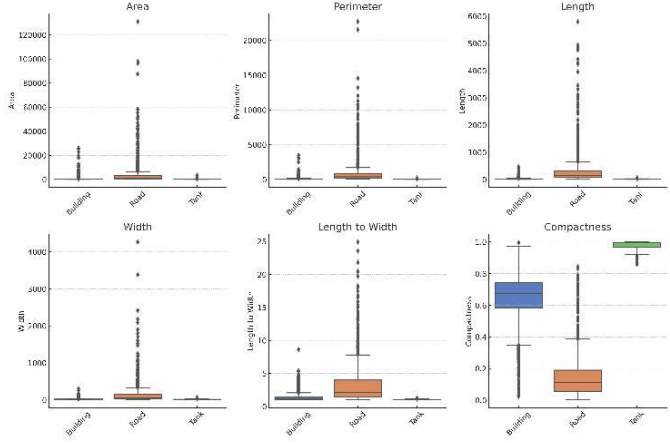

**Figure A1(a):** Boxplots of geometric properties of infrastructure training data measured in meters.

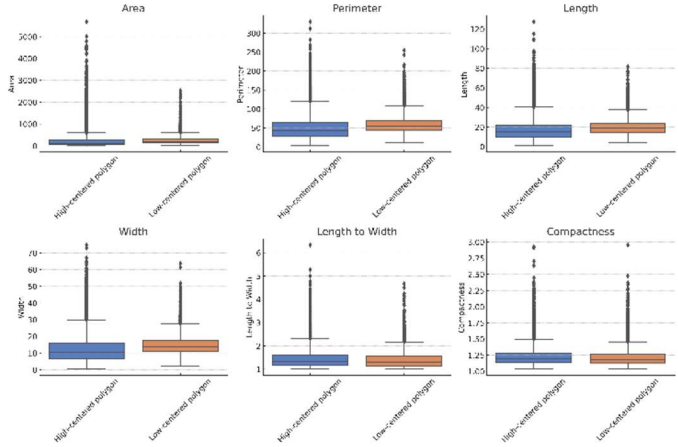

**Figure A1(b):** Boxplots of geometric properties of ice-wedge polygon training data measured in meters.

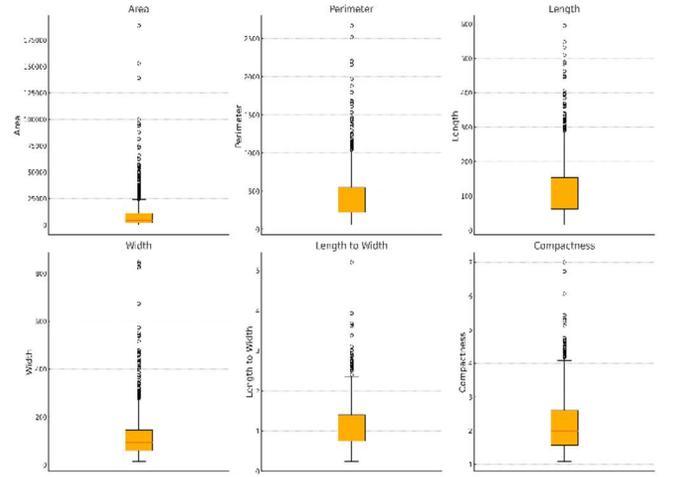

**Figure A1(c):** Boxplots of geometric properties of retrogressive thaw slump, training data measured in meters.

We present detailed results (table A1-AV) of all the ablation experiments carried out in this appendix.

TABLE AI
ViT Size Validation metrics for RTS

| Size | Pix. Ac | Prec. | Rec. | F1 | mIoU |
|------|---------|-------|------|-----|------|
| Base | 0.9406 | 0.8217 | 0.7973 | 0.8086 | 0.7091 |
|      | ±0.0023 | ±0.0077 | ±0.0091 | ±0.0026 | ±0.0033 |
| Large | 0.9222 | 0.8278 | 0.7463 | 0.7781 | 0.6728 |
|       | ±0.0091 | ±0.0071 | ±0.0189 | ±0.0124 | ±0.0147 |
| Huge | 0.9401 | 0.8320 | 0.7932 | 0.8110 | 0.7117 |
|      | ±0.0006 | ±0.0039 | ±0.0016 | ±0.0025 | ±0.0029 |

TABLE AII
ViT Size Validation metrics for Infrastructure

| Size | Pix. Ac | Prec. | Rec. | F1 | mIoU |
|------|---------|-------|------|-----|------|
| Base | 0.9507 | 0.7888 | 0.8027 | 0.7939 | 0.6751 |
|      | ±0.0003 | ±0.0132 | ±0.0116 | ±0.0072 | ±0.0082 |
| Large | 0.9463 | 0.8204 | 0.7491 | 0.7674 | 0.6437 |
|       | ±0.0001 | ±0.0028 | ±0.0043 | ±0.0035 | ±0.0036 |
| Huge | 0.9476 | 0.8358 | 0.7724 | 0.7947 | 0.6755 |
|      | ±0.0018 | ±0.0172 | ±0.0178 | ±0.0172 | ±0.0202 |

TABLE AIII
ViT Size Validation metrics for IWP

| Size | mAP | mAP$_{50}$ | mAP$_{75}$ | APs | APm | APl |
|------|-----|------|------|-----|-----|-----|
| Base | 0.2506 | 0.4820 | 0.2474 | 0.1944 | 0.4484 | 0.5754 |
|      | ±0.0034 | ±0.0037 | ±0.0061 | ±0.0027 | ±0.0075 | ±0.0132 |
| Large | 0.2721 | 0.5077 | 0.2764 | 0.2158 | 0.4724 | 0.6161 |
|       | ±0.0050 | ±0.0070 | ±0.0076 | ±0.005 | ±0.0030 | ±0.0071 |
| Huge | 0.2688 | 0.5020 | 0.2766 | 0.2103 | 0.4651 | 0.6005 |
|      | ±0.0036 | ±0.0057 | ±0.0055 | ±0.0039 | ±0.0065 | ±0.0088 |

TABLE A IV
Location Embedding Fusion Validation metrics for IWP Experiments

| Feature | Encoder | Place. | Merge Strategy | mAP | mAP$_{50}$ | mAP$_{75}$ | APs | APm | APl |
|---------|---------|--------|----------------|-----|------|------|-----|-----|-----|
| IWP | L10 | Post | Add | 0.2445±0.0035 | 0.4751±0.0054 | 0.2387±0.0089 | 0.1885±0.0046 | 0.4344±0.0013 | 0.5479±0.0040 |
|     |     |      | Norm Add | 0.2514±0.0031 | 0.4847±0.0052 | 0.2455±0.0019 | 0.1976±0.0028 | 0.4353±0.0053 | 0.5610±0.0045 |
|     |     |      | Concat | 0.2472±0.0025 | 0.4808±0.0052 | 0.2449±0.0015 | 0.1950±0.0017 | 0.4309±0.0055 | 0.5453±0.0131 |
|     |     |      | Norm Concat | 0.2492±0.0019 | 0.4807±0.0014 | 0.2476±0.0030 | 0.1963±0.0018 | 0.4342±0.0037 | 0.5689±0.0081 |
|     |     |      | Concat Norm | 0.2497±0.0043 | 0.4871±0.0052 | 0.2451±0.0047 | 0.1951±0.0041 | 0.4370±0.0055 | 0.5517±0.0083 |
|     |     |      | Projection Concat | 0.2531±0.0031 | 0.4854±0.0043 | 0.2530±0.0049 | 0.1978±0.0017 | 0.4455±0.0094 | 0.5989±0.0150 |
|     |     |      | Projection Add | 0.2508±0.0030 | 0.4813±0.0026 | 0.2500±0.0048 | 0.1964±0.0026 | 0.4422±0.0032 | 0.5680±0.0153 |
|     |     |      | Cross Attention | 0.2488±0.0020 | 0.4844±0.0034 | 0.2436±0.0030 | 0.1959±0.0019 | 0.4412±0.0034 | 0.5619±0.0090 |
| IWP | L10 | Pre | Concat | 0.2534±0.0001 | 0.4849±0.0031 | 0.2547±0.0028 | 0.1987±0.0004 | 0.4465±0.0012 | 0.5801±0.0066 |
|     |     |     | Norm Concat | 0.2521±0.0019 | 0.4823±0.0030 | 0.2505±0.0039 | 0.1980±0.0033 | 0.4466±0.0051 | 0.5909±0.0114 |
|     |     |     | Concat Norm | 0.2446±0.0042 | 0.4718±0.0054 | 0.2433±0.0081 | 0.1903±0.0032 | 0.4358±0.0068 | 0.5625±0.0140 |
|     |     |     | Projection Concat | 0.2448±0.0012 | 0.4719±0.0027 | 0.2412±0.0031 | 0.1904±0.0015 | 0.4362±0.0008 | 0.5793±0.0114 |
|     |     |     | Projection Add | 0.2532±0.0007 | 0.4851±0.0001 | 0.2519±0.0017 | 0.1986±0.0002 | 0.4428±0.0032 | 0.5664±0.0023 |
|     |     |     | Cross Attention | 0.2496±0.0002 | 0.4821±0.0004 | 0.2453±0.0059 | 0.1973±0.0003 | 0.4372±0.0043 | 0.5835±0.0072 |

TABLE A IV (CONTINUED)

| Feature | Encoder | Place. | Merge Strategy | mAP | mAP$_{50}$ | mAP$_{75}$ | APs | APm | APl |
|---|---|---|---|---|---|---|---|---|---|
| IWP | L40 | Post | Add | 0.2444±0.0044 | 0.4789±0.0053 | 0.2387±0.0061 | 0.1886±0.0038 | 0.4382±0.0070 | 0.5735±0.0144 |
| | | | Norm Add | 0.2498±0.0031 | 0.4827±0.0034 | 0.2492±0.0046 | 0.1955±0.0022 | 0.4422±0.0060 | 0.5767±0.0185 |
| | | | Concat | 0.2441±0.0033 | 0.4773±0.0044 | 0.2364±0.0037 | 0.1878±0.0031 | 0.4384±0.0054 | 0.5782±0.0263 |
| | | | Norm Concat | 0.2545±0.0027 | 0.4907±0.0020 | 0.2519±0.0050 | 0.2004±0.0021 | 0.4445±0.0042 | 0.5663±0.0119 |
| | | | Concat Norm | 0.2436±0.0033 | 0.4773±0.0037 | 0.2390±0.0052 | 0.1906±0.0019 | 0.4316±0.0098 | 0.5686±0.0183 |
| | | | Projection Concat | 0.2495±0.0028 | 0.4806±0.0041 | 0.2472±0.0065 | 0.1985±0.0015 | 0.4327±0.0023 | 0.5741±0.0080 |
| | | | Projection Add | 0.2508±0.0049 | 0.4825±0.0057 | 0.2507±0.0084 | 0.1986±0.0059 | 0.4411±0.0050 | 0.5723±0.0085 |
| | | | Cross Attention | 0.2475±0.0025 | 0.4754±0.0023 | 0.2439±0.0055 | 0.1938±0.0032 | 0.4406±0.0060 | 0.5769±0.0076 |
| IWP | L40 | Pre | Concat | 0.2451±0.0001 | 0.4840±0.0001 | 0.2363±0.0001 | 0.1915±0.0001 | 0.4421±0.0001 | 0.5756±0.0001 |
| | | | Norm Concat | 0.2540±0.0001 | 0.4841±0.0001 | 0.2570±0.0001 | 0.1988±0.0001 | 0.4389±0.0001 | 0.5803±0.0001 |
| | | | Concat Norm | 0.2397±0.0006 | 0.4718±0.0031 | 0.2294±0.0044 | 0.1830±0.0031 | 0.4345±0.0038 | 0.5551±0.0054 |
| | | | Projection Concat | 0.2476±0.0029 | 0.4774±0.0035 | 0.2444±0.0032 | 0.1943±0.0034 | 0.4346±0.0081 | 0.5894±0.0095 |
| | | | Projection Add | 0.2518±0.0022 | 0.4854±0.0027 | 0.2512±0.0002 | 0.1964±0.0013 | 0.4431±0.0077 | 0.5645±0.0219 |
| | | | Cross Attention | 0.2548±0.0012 | 0.4917±0.0025 | 0.2533±0.0019 | 0.2001±0.0017 | 0.4446±0.0007 | 0.5750±0.0114 |

TABLE A V
LOCATION EMBEDDING FUSION VALIDATION METRICS FOR RTS AND INFRASTRUCTURE EXPERIMENTS

| Feature | Encoder | Place. | Merge Strategy | Pixel Accuracy | Precision | Recall | F1 score | mIoU |
|---|---|---|---|---|---|---|---|---|
| Infra | L10 | Post | Add | 0.9639±0.0002 | 0.8706±0.0024 | 0.8857±0.0039 | 0.8780±0.0014 | 0.7894±0.0022 |
| | | | Norm Add | 0.9640±0.0001 | 0.8656±0.0017 | 0.8876±0.0016 | 0.8764±0.0016 | 0.7869±0.0024 |
| | | | Concat | 0.9644±0.0001 | 0.8511±0.0034 | 0.8962±0.0020 | 0.8726±0.0016 | 0.7814±0.0023 |
| | | | Norm Concat | 0.9653±0.0001 | 0.8597±0.0026 | 0.8921±0.0014 | 0.8753±0.0009 | 0.7854±0.0012 |
| | | | Concat Norm | 0.9638±0.0020 | 0.8250±0.0060 | 0.9020±0.0020 | 0.8606±0.0026 | 0.7641±0.0036 |
| | | | Projection Concat | 0.9659±0.0001 | 0.8740±0.0008 | 0.9074±0.0005 | 0.8901±0.0006 | 0.8080±0.0009 |
| | | | Projection Add | 0.9638±0.0002 | 0.8576±0.0049 | 0.8962±0.0010 | 0.8755±0.0029 | 0.7856±0.0043 |
| | | | Cross Attention | 0.9648±0.0001 | 0.8627±0.0034 | 0.8949±0.0014 | 0.8782±0.0017 | 0.7895±0.0021 |
| Infra | L10 | Pre | Concat | 0.9653±0.0001 | 0.8866±0.0034 | 0.8778±0.0023 | 0.8817±0.0016 | 0.7946±0.0024 |
| | | | Norm Concat | 0.9655±0.0001 | 0.8808±0.0063 | 0.8856±0.0015 | 0.8831±0.0028 | 0.7968±0.0043 |
| | | | Concat Norm | 0.9655±0.0001 | 0.8765±0.0056 | 0.8763±0.0021 | 0.8762±0.0030 | 0.7865±0.0044 |
| | | | Projection Concat | 0.9652±0.0001 | 0.8761±0.0013 | 0.8882±0.0012 | 0.8821±0.0007 | 0.7955±0.0011 |
| | | | Projection Add | 0.9646±0.0001 | 0.8610±0.0040 | 0.8904±0.0026 | 0.8751±0.0027 | 0.7849±0.0040 |
| | | | Cross Attention | 0.9656±0.0001 | 0.8848±0.0004 | 0.8881±0.0002 | 0.8863±0.0002 | 0.8019±0.0003 |
| Infra | L40 | Post | Add | 0.9625±0.0001 | 0.8695±0.0035 | 0.8821±0.0039 | 0.8756±0.0019 | 0.7858±0.0030 |
| | | | Norm Add | 0.9637±0.0001 | 0.8585±0.0028 | 0.8911±0.0010 | 0.8742±0.0015 | 0.7836±0.0022 |
| | | | Concat | 0.9649±0.0001 | 0.8705±0.0020 | 0.8960±0.0021 | 0.8828±0.0012 | 0.7966±0.0019 |
| | | | Norm Concat | 0.9651±0.0001 | 0.8666±0.0089 | 0.8894±0.0024 | 0.8776±0.0050 | 0.7889±0.0074 |
| | | | Concat Norm | 0.9645±0.0003 | 0.8523±0.0057 | 0.8954±0.0052 | 0.8727±0.0018 | 0.7813±0.0026 |
| | | | Projection Concat | 0.9657±0.0001 | 0.8633±0.0050 | 0.8931±0.0018 | 0.8776±0.0021 | 0.7888±0.0031 |
| | | | Projection Add | 0.9638±0.0001 | 0.8573±0.0024 | 0.8876±0.0014 | 0.8719±0.0015 | 0.7804±0.0022 |
| | | | Cross Attention | 0.9655±0.0001 | 0.8719±0.0014 | 0.8931±0.0010 | 0.8822±0.0010 | 0.7955±0.0015 |
| Infra | L40 | Pre | Concat | 0.9645±0.0001 | 0.8475±0.0035 | 0.8903±0.0008 | 0.8679±0.0021 | 0.7746±0.0029 |
| | | | Norm Concat | 0.9655±0.0001 | 0.8708±0.0013 | 0.8897±0.0012 | 0.8800±0.0007 | 0.7924±0.0010 |
| | | | Concat Norm | 0.9643±0.0001 | 0.8496±0.0015 | 0.8911±0.0016 | 0.8693±0.0008 | 0.7768±0.0011 |
| | | | Projection Concat | 0.9642±0.0001 | 0.8583±0.0102 | 0.8943±0.0025 | 0.8755±0.0049 | 0.7853±0.0073 |
| | | | Projection Add | 0.9646±0.0001 | 0.8659±0.0028 | 0.8884±0.0031 | 0.8769±0.0018 | 0.7875±0.0027 |
| | | | Cross Attention | 0.9657±0.9645 | 0.8592±0.8475 | 0.901±0.8903 | 0.8787±0.8679 | 0.7904±0.7746 |
| RTS | L10 | Post | Add | 0.9463±0.0040 | 0.8136±0.0243 | 0.8249±0.0220 | 0.8171±0.0075 | 0.7199±0.0087 |
| | | | Norm Add | 0.9469±0.0011 | 0.8347±0.0024 | 0.8180±0.0046 | 0.8260±0.0022 | 0.7300±0.0028 |
| | | | Concat | 0.0809±0.0001 | 0.5000±0.0001 | 0.5405±0.0001 | 0.0749±0.0001 | 0.0405±0.0001 |
| | | | Norm Concat | 0.5469±0.0635 | 0.7108±0.0269 | 0.5638±0.0072 | 0.4602±0.0420 | 0.3260±0.0417 |
| | | | Concat Norm | 0.0812±0.0002 | 0.5002±0.0001 | 0.5401±0.0010 | 0.0752±0.0003 | 0.0406±0.0001 |
| | | | Projection Concat | 0.9481±0.0007 | 0.8404±0.0020 | 0.8215±0.0029 | 0.8306±0.0017 | 0.7355±0.0021 |
| | | | Projection Add | 0.9481±0.0012 | 0.8393±0.0018 | 0.8220±0.0052 | 0.8303±0.0024 | 0.7352±0.0030 |
| | | | Cross Attention | 0.9503±0.0008 | 0.8283±0.0029 | 0.8345±0.0037 | 0.8313±0.0023 | 0.7369±0.0028 |
| RTS | L10 | Pre | Concat | 0.9546±0.0007 | 0.8123±0.0058 | 0.8646±0.0065 | 0.8358±0.0019 | 0.7430±0.0022 |
| | | | Norm Concat | 0.9543±0.0008 | 0.8402±0.0036 | 0.8491±0.0052 | 0.8445±0.0016 | 0.7533±0.0020 |
| | | | Concat Norm | 0.9557±0.0008 | 0.8218±0.0055 | 0.8650±0.0062 | 0.8416±0.0023 | 0.7501±0.0029 |
| | | | Projection Concat | 0.9543±0.0008 | 0.8344±0.0036 | 0.8514±0.0044 | 0.8426±0.0022 | 0.7510±0.0028 |
| | | | Projection Add | 0.9501±0.0017 | 0.8438±0.0027 | 0.8289±0.0073 | 0.8361±0.0039 | 0.7424±0.0050 |
| | | | Cross Attention | 0.9528±0.0015 | 0.8462±0.0026 | 0.8398±0.0064 | 0.8429±0.0035 | 0.7510±0.0045 |
| RTS | L40 | Post | Add | 0.9500±0.0009 | 0.7676±0.0309 | 0.8687±0.0218 | 0.8057±0.0145 | 0.7085±0.0155 |
| | | | Norm Add | 0.9547±0.0004 | 0.8153±0.0058 | 0.8633±0.0032 | 0.8371±0.0026 | 0.7446±0.0031 |
| | | | Concat | 0.7462±0.3175 | 0.6912±0.0980 | 0.7104±0.1189 | 0.5952±0.2423 | 0.5001±0.2224 |
| | | | Norm Concat | 0.9493±0.0022 | 0.8654±0.0087 | 0.8210±0.0101 | 0.8410±0.0025 | 0.7481±0.0034 |
| | | | Concat Norm | 0.2714±0.1542 | 0.5890±0.0698 | 0.5458±0.0063 | 0.2475±0.1310 | 0.1547±0.0932 |
| | | | Projection Concat | 0.9515±0.0010 | 0.8438±0.0015 | 0.8347±0.0040 | 0.8391±0.0026 | 0.7463±0.0034 |
| | | | Projection Add | 0.9553±0.0009 | 0.8417±0.0021 | 0.8530±0.0056 | 0.8472±0.0016 | 0.7567±0.0021 |
| | | | Cross Attention | 0.9529±0.0015 | 0.8351±0.0020 | 0.8447±0.0079 | 0.8397±0.0032 | 0.7473±0.0041 |
| RTS | L40 | Pre | Concat | 0.9554±0.0002 | 0.8488±0.0017 | 0.8507±0.0002 | 0.8498±0.0010 | 0.7599±0.0012 |
| | | | Norm Concat | 0.9546±0.0001 | 0.8490±0.0001 | 0.8465±0.0001 | 0.8478±0.0001 | 0.7573±0.0001 |
| | | | Concat Norm | 0.9565±0.0004 | 0.8343±0.0010 | 0.8627±0.0027 | 0.8478±0.0007 | 0.7577±0.0009 |
| | | | Projection Concat | 0.9568±0.0001 | 0.8345±0.0001 | 0.8642±0.0001 | 0.8486±0.0001 | 0.7587±0.0001 |
| | | | Projection Add | 0.9537±0.0001 | 0.8355±0.0010 | 0.8478±0.0001 | 0.8415±0.0001 | 0.7496±0.0001 |
| | | | Cross Attention | 0.9552±0.0001 | 0.8377±0.0028 | 0.8544±0.0017 | 0.8458±0.0006 | 0.7550±0.0008 |

## REFERENCES

[1] S. Gartler et al., "A transdisciplinary, comparative analysis reveals key risks from Arctic permafrost thaw," Commun. Earth Environ., vol. 6, no. 1, pp. 1–20, Jan. 2025, doi: 10.1038/s43247-024-01883-w.

[2] T. L. Spero, N. L. Briggs, and L. Boldrick, "Environmental Impacts from Projected Permafrost Thaw in Alaska: Defining Knowledge Gaps, Data Needs, and Research Priorities," Apr. 2025, doi: 10.1175/WCAS-D-24-0150.1.

[3] Alaska Native Tribal Health Consortium, "Unmet Needs of Environmentally Threatened Alaska Native Villages: Assessment and Recommendations," ANTHC, AL, 2024.

[4] M. Rantanen et al., "The Arctic has warmed nearly four times faster than the globe since 1979," Commun. Earth Environ., vol. 3, no. 1, pp. 1–10, Aug. 2022, doi: 10.1038/s43247-022-00498-3.

[5] S. V. Kokelj, T. C. Lantz, J. Tunnicliffe, R. Segal, and D. Lacelle, "Climate-driven thaw of permafrost preserved glacial landscapes, northwestern Canada," Geology, vol. 45, no. 4, pp. 371–374, Apr. 2017, doi: 10.1130/G38626.1.

[6] D. M. Nielsen et al., "Increase in Arctic coastal erosion and its sensitivity to warming in the twenty-first century," Nat. Clim. Change, vol. 12, no. 3, pp. 263–270, Mar. 2022, doi: 10.1038/s41558-022-01281-0.

[7] J. Hjort, D. Streletskiy, G. Doré, Q. Wu, K. Bjella, and M. Luoto, "Impacts of permafrost degradation on infrastructure," Nat. Rev. Earth Environ., vol. 3, no. 1, pp. 24–38, Jan. 2022, doi: 10.1038/s43017-021-00247-8.

[8] E. A. G. Schuur et al., "Permafrost and Climate Change: Carbon Cycle Feedbacks From the Warming Arctic," 2022, doi: https://doi.org/10.1146/annurev-environ-012220-011847.

[9] A. Bartsch et al., "Towards long-term records of rain-on-snow events across the Arctic from satellite data," The Cryosphere, vol. 17, no. 2, pp. 889–915, Feb. 2023, doi: 10.5194/tc-17-889-2023.

[10] A. K. Liljedahl, C. Witharana, and E. Manos, "The capillaries of the Arctic tundra," Nat. Water, vol. 2, no. 7, pp. 611–614, 2024.

[11] M. Udawalpola, A. Hasan, A. Liljedahl, A. Soliman, J. Terstriep, and C. Witharana, "An Optimal GeoAI Workflow for Pan-Arctic Permafrost Feature Detection from High-Resolution Satellite Imagery," Photogramm. Eng. Remote Sens., vol. 88, pp. 181–188, Mar. 2022, doi: 10.14358/PERS.21-00059R2.

[12] A. Hasan, M. Udawalpola, A. Liljedahl, and C. Witharana, "Use of commercial satellite imagery to monitor changing arctic polygonal tundra," Photogramm. Eng. Remote Sens., vol. 88, no. 4, pp. 255–262, 2022.

[13] C. Witharana et al., "An Object-Based Approach for Mapping Tundra Ice-Wedge Polygon Troughs from Very High Spatial Resolution Optical Satellite Imagery," Remote Sens., vol. 13, no. 4, Art. no. 4, Jan. 2021, doi: 10.3390/rs13040558.

[14] S. Barth, I. Nitze, B. Juhls, A. Runge, and G. Grosse, "Rapid Changes in Retrogressive Thaw Slump Dynamics in the Russian High Arctic Based on Very High-Resolution Remote Sensing," Geophys. Res. Lett., vol. 52, no. 7, p. e2024GL113022, 2025, doi: 10.1029/2024GL113022.

[15] Y. Yang et al., "A Collaborative and Scalable Geospatial Data Set for Arctic Retrogressive Thaw Slumps with Data Standards," Sci. Data, vol. 12, no. 1, p. 18, Jan. 2025, doi: 10.1038/s41597-025-04372-7.

[16] E. Manos, C. Witharana, A. S. Perera, and A. K. Liljedahl, "A multi-objective comparison of CNN architectures in Arctic human-built infrastructure mapping from sub-meter resolution satellite imagery," Int. J. Remote Sens., vol. 44, no. 24, pp. 7670–7705, Dec. 2023, doi: 10.1080/01431161.2023.2287563.

[17] E. Manos, C. Witharana, and A. K. Liljedahl, "Permafrost thaw-related infrastructure damage costs in Alaska are projected to double under medium and high emission scenarios," Commun. Earth Environ., vol. 6, no. 1, pp. 1–11, Mar. 2025, doi: 10.1038/s43247-025-02191-7.

[18] A. Dosovitskiy et al., "An Image is Worth 16x16 Words: Transformers for Image Recognition at Scale," in 9th International Conference on Learning Representations, ICLR 2021, Virtual Event, Austria, May 3-7, 2021, OpenReview.net, 2021. Accessed: Nov. 15, 2024. [Online]. Available: https://openreview.net/forum?id=YicbFdNTTy

[19] N. Ebert, D. Stricker, and O. Wasenmüller, "PLG-ViT: Vision Transformer with Parallel Local and Global Self-Attention," Sensors, vol. 23, no. 7, p. 3447, Mar. 2023, doi: 10.3390/s23073447.

[20] Y. Wang, C. M. Albrecht, and X. X. Zhu, "Self-Supervised Vision Transformers for Joint SAR-Optical Representation Learning," in IGARSS 2022 - 2022 IEEE International Geoscience and Remote Sensing Symposium, Jul. 2022, pp. 139–142. doi: 10.1109/IGARSS46834.2022.9883983.

[21] Y. Cong et al., "SatMAE: pre-training transformers for temporal and multi-spectral satellite imagery," in Proceedings of the 36th International Conference on Neural Information Processing Systems, in NIPS '22. Red Hook, NY, USA: Curran Associates Inc., Nov. 2022, pp. 197–211.

[22] X. Huang and L. Zhang, "Morphological Building/Shadow Index for Building Extraction From High-Resolution Imagery Over Urban Areas," IEEE J. Sel. Top. Appl. Earth Obs. Remote Sens., vol. 5, no. 1, pp. 161–172, Feb. 2012, doi: 10.1109/JSTARS.2011.2168195.

[23] I. Demir et al., "DeepGlobe 2018: A Challenge to Parse the Earth through Satellite Images," in 2018 IEEE/CVF Conference on Computer Vision and Pattern Recognition Workshops (CVPRW), Salt Lake City, UT, USA: IEEE, Jun. 2018, pp. 172–17209. doi: 10.1109/CVPRW.2018.00031.

[24] V. V. Cepeda, G. K. Nayak, and M. Shah, "GeoCLIP: Clip-Inspired Alignment between Locations and Images for Effective Worldwide Geo-localization," Nov. 21, 2023, arXiv: arXiv:2309.16020. doi: 10.48550/arXiv.2309.16020.

[25] A. Radford et al., "Learning Transferable Visual Models From Natural Language Supervision," in Proceedings of the 38th International Conference on Machine Learning, PMLR, Jul. 2021, pp. 8748–8763. Accessed: Jan. 02, 2025. [Online]. Available: https://proceedings.mlr.press/v139/radford21a.html

[26] S. Tucker, "A systematic review of geospatial location embedding approaches in large language models: A path to spatial AI systems," Jan. 12, 2024, arXiv: arXiv:2401.10279. doi: 10.48550/arXiv.2401.10279.

[27] K. Klemmer, E. Rolf, C. Robinson, L. Mackey, and M. Rußwurm, "SatCLIP: Global, General-Purpose Location Embeddings with Satellite Imagery," Apr. 12, 2024, arXiv: arXiv:2311.17179. doi: 10.48550/arXiv.2311.17179.

[28] G. M. Foody and A. Mathur, "A relative evaluation of multiclass image classification by support vector machines," IEEE Trans. Geosci. Remote Sens., vol. 42, no. 6, pp. 1335–1343, Jun. 2004, doi: 10.1109/TGRS.2004.827257.

[29] Y. Tarabalka, Jó. A. Benediktsson, and J. Chanussot, "Spectral–Spatial Classification of Hyperspectral Imagery Based on Partitional Clustering Techniques," IEEE Trans. Geosci. Remote Sens., vol. 47, no. 8, pp. 2973–2987, Aug. 2009, doi: 10.1109/TGRS.2009.2016214.

[30] W. Li, S. Prasad, J. E. Fowler, and L. M. Bruce, "Locality-Preserving Dimensionality Reduction and Classification for Hyperspectral Image Analysis," IEEE Trans. Geosci. Remote Sens., vol. 50, no. 4, pp. 1185–1198, Apr. 2012, doi: 10.1109/TGRS.2011.2165957.

[31] F. Melgani and L. Bruzzone, "Classification of hyperspectral remote sensing images with support vector machines," IEEE Trans. Geosci. Remote Sens., vol. 42, no. 8, pp. 1778–1790, Aug. 2004, doi: 10.1109/TGRS.2004.831865.

[32] J. A. Gualtieri and S. Chettri, "Support vector machines for classification of hyperspectral data," in IGARSS 2000. IEEE 2000 International Geoscience and Remote Sensing Symposium. Taking the Pulse of the Planet: The Role of Remote Sensing in Managing the Environment. Proceedings (Cat. No.00CH37120), Jul. 2000, pp. 813–815 vol.2. doi: 10.1109/IGARSS.2000.861712.

[33] G. Mountrakis, J. Im, and C. Ogole, "Support vector machines in remote sensing: A review," ISPRS J. Photogramm. Remote Sens., vol. 66, no. 3, pp. 247–259, May 2011, doi: 10.1016/j.isprsjprs.2010.11.001.

[34] P. M. Atkinson and A. R. L. Tatnall, "Introduction Neural networks in remote sensing," Int. J. Remote Sens., vol. 18, no. 4, pp. 699–709, Mar. 1997, doi: 10.1080/014311697218700.

[35] L. Bruzzone and D. F. Prieto, "A technique for the selection of kernel-function parameters in RBF neural networks for classification of remote-sensing images," IEEE Trans. Geosci. Remote Sens., vol. 37, no. 2, pp. 1179–1184, Mar. 1999, doi: 10.1109/36.752239.

[36] G. E. Hinton and R. R. Salakhutdinov, "Reducing the Dimensionality of Data with Neural Networks," Science, vol. 313, no. 5786, pp. 504–507, Jul. 2006, doi: 10.1126/science.1127647.

[37] L. Ma, Y. Liu, X. Zhang, Y. Ye, G. Yin, and B. A. Johnson, "Deep learning in remote sensing applications: A meta-analysis and review," ISPRS J. Photogramm. Remote Sens., vol. 152, pp. 166–177, Jun. 2019, doi: 10.1016/j.isprsjprs.2019.04.015.

[38] W. Hu, Y. Huang, L. Wei, F. Zhang, and H. Li, "Deep Convolutional Neural Networks for Hyperspectral Image Classification," J. Sens., vol. 2015, no. 1, p. 258619, 2015, doi: 10.1155/2015/258619.

[39] X. X. Zhu *et al.*, "Deep Learning in Remote Sensing: A Comprehensive Review and List of Resources," *IEEE Geosci. Remote Sens. Mag.*, vol. 5, no. 4, pp. 8–36, Dec. 2017, doi: 10.1109/MGRS.2017.2762307.

[40] J. Long, E. Shelhamer, and T. Darrell, "Fully convolutional networks for semantic segmentation," in *2015 IEEE Conference on Computer Vision and Pattern Recognition (CVPR)*, Jun. 2015, pp. 3431–3440. doi: 10.1109/CVPR.2015.7298965.

[41] E. Xie, W. Wang, Z. Yu, A. Anandkumar, J. M. Alvarez, and P. Luo, "SegFormer: Simple and Efficient Design for Semantic Segmentation with Transformers," in *Advances in Neural Information Processing Systems*, Curran Associates, Inc., 2021, pp. 12077–12090. Accessed: Jan. 03, 2025. [Online]. Available: https://proceedings.neurips.cc/paper/2021/hash/64f1f27bf1b4ec22924fd0acb550c235-Abstract.html

[42] J. Shotton, M. Johnson, and R. Cipolla, "Semantic texton forests for image categorization and segmentation," in *2008 IEEE Conference on Computer Vision and Pattern Recognition*, Jun. 2008, pp. 1–8. doi: 10.1109/CVPR.2008.4587503.

[43] O. Ronneberger, P. Fischer, and T. Brox, "U-Net: Convolutional Networks for Biomedical Image Segmentation," in *Medical Image Computing and Computer-Assisted Intervention – MICCAI 2015*, N. Navab, J. Hornegger, W. M. Wells, and A. F. Frangi, Eds., Cham: Springer International Publishing, 2015, pp. 234–241. doi: 10.1007/978-3-319-24574-4_28.

[44] K. He, G. Gkioxari, P. Dollár, and R. Girshick, "Mask r-cnn," presented at the Proceedings of the IEEE international conference on computer vision, 2017, pp. 2961–2969.

[45] Z. Tian, C. Shen, H. Chen, and T. He, "FCOS: Fully Convolutional One-Stage Object Detection," in *2019 IEEE/CVF International Conference on Computer Vision (ICCV)*, Oct. 2019, pp. 9626–9635. doi: 10.1109/ICCV.2019.00972.

[46] Y. Lee and J. Park, "CenterMask: Real-Time Anchor-Free Instance Segmentation," in *2020 IEEE/CVF Conference on Computer Vision and Pattern Recognition (CVPR)*, Jun. 2020, pp. 13903–13912. doi: 10.1109/CVPR42600.2020.01392.

[47] B. Cheng, I. Misra, A. G. Schwing, A. Kirillov, and R. Girdhar, "Masked-Attention Mask Transformer for Universal Image Segmentation," presented at the Proceedings of the IEEE/CVF Conference on Computer Vision and Pattern Recognition, 2022, pp. 1290–1299. Accessed: Jan. 03, 2025. [Online]. Available: https://openaccess.thecvf.com/content/CVPR2022/html/Cheng_Masked-Attention_Mask_Transformer_for_Universal_Image_Segmentation_CVPR_2022_paper.html

[48] Z. Liu *et al.*, "Swin Transformer: Hierarchical Vision Transformer using Shifted Windows," in *2021 IEEE/CVF International Conference on Computer Vision (ICCV)*, Oct. 2021, pp. 9992–10002. doi: 10.1109/ICCV48922.2021.00986.

[49] H. Touvron, M. Cord, M. Douze, F. Massa, A. Sablayrolles, and H. Jegou, "Training data-efficient image transformers & distillation through attention," in *Proceedings of the 38th International Conference on Machine Learning*, PMLR, Jul. 2021, pp. 10347–10357. Accessed: Mar. 05, 2025. [Online]. Available: https://proceedings.mlr.press/v139/touvron21a.html

[50] R. Girshick, J. Donahue, T. Darrell, and J. Malik, "Rich Feature Hierarchies for Accurate Object Detection and Semantic Segmentation," in *2014 IEEE Conference on Computer Vision and Pattern Recognition*, Jun. 2014, pp. 580–587. doi: 10.1109/CVPR.2014.81.

[51] J. Deng, W. Dong, R. Socher, L.-J. Li, K. Li, and L. Fei-Fei, "ImageNet: A large-scale hierarchical image database," in *2009 IEEE Conference on Computer Vision and Pattern Recognition*, Jun. 2009, pp. 248–255. doi: 10.1109/CVPR.2009.5206848.

[52] A. Krizhevsky, I. Sutskever, and G. E. Hinton, "ImageNet Classification with Deep Convolutional Neural Networks," in *Advances in Neural Information Processing Systems*, Curran Associates, Inc., 2012. Accessed: Apr. 07, 2025. [Online]. Available: https://proceedings.neurips.cc/paper_files/paper/2012/hash/c399862d3b9d6b76c8436e924a68c45b-Abstract.html

[53] K. Simonyan and A. Zisserman, "Very Deep Convolutional Networks for Large-Scale Image Recognition," Apr. 10, 2015, *arXiv*: arXiv:1409.1556. doi: 10.48550/arXiv.1409.1556.

[54] K. He, X. Zhang, S. Ren, and J. Sun, "Deep Residual Learning for Image Recognition," in *2016 IEEE Conference on Computer Vision and Pattern Recognition (CVPR)*, Jun. 2016, pp. 770–778. doi: 10.1109/CVPR.2016.90.

[55] H. Touvron, M. Cord, M. Douze, F. Massa, A. Sablayrolles, and H. Jegou, "Training data-efficient image transformers & distillation through attention," in *Proceedings of the 38th International Conference on Machine Learning*, PMLR, Jul. 2021, pp. 10347–10357. Accessed: Mar. 05, 2025. [Online]. Available: https://proceedings.mlr.press/v139/touvron21a.html

[56] Y. Li, H. Mao, R. Girshick, and K. He, "Exploring Plain Vision Transformer Backbones for Object Detection," Jun. 10, 2022, *arXiv*: arXiv:2203.16527. doi: 10.48550/arXiv.2203.16527.

[57] J. Chen *et al.*, "TransUNet: Rethinking the U-Net architecture design for medical image segmentation through the lens of transformers," *Med. Image Anal.*, vol. 97, p. 103280, Oct. 2024, doi: 10.1016/j.media.2024.103280.

[58] C. Witharana *et al.*, "Automated Detection of Retrogressive Thaw Slumps in the High Arctic Using High-Resolution Satellite Imagery," *Remote Sens.*, vol. 14, no. 17, Art. no. 17, Jan. 2022, doi: 10.3390/rs14174132.

[59] E. Manos, C. Witharana, M. R. Udawalpola, A. Hasan, and A. K. Liljedahl, "Convolutional Neural Networks for Automated Built Infrastructure Detection in the Arctic Using Sub-Meter Spatial Resolution Satellite Imagery," *Remote Sens.*, vol. 14, no. 11, Art. no. 11, Jan. 2022, doi: 10.3390/rs14112719.

[60] A. S. Perera, C. Witharana, E. Manos, and A. K. Liljedahl, "Hyperparameter Optimization for Large-Scale Remote Sensing Image Analysis Tasks: A Case Study Based on Permafrost Landform Detection Using Deep Learning," *IEEE Access*, vol. 12, pp. 43062–43077, 2024.

[61] Yuxin Wu, Alexander Kirillov, Wan-Yen Lo, and Ross Girshick, *Detectron2*. (2019). [Online]. Available: https://github.com/facebookresearch/detectron2

[62] D. Stanzione, J. West, R. T. Evans, T. Minyard, O. Ghattas, and D. K. Panda, "Frontera: The Evolution of Leadership Computing at the National Science Foundation," in *Practice and Experience in Advanced Research Computing*, in PEARC '20. New York, NY, USA: Association for Computing Machinery, Jul. 2020, pp. 106–111. doi: 10.1145/3311790.3396656.

[63] "Lonestar6." Accessed: Jan. 07, 2025. [Online]. Available: http://tacc.utexas.edu/systems/lonestar6/

[64] "Google Cloud Documentation," Google Cloud. Accessed: Jan. 07, 2025. [Online]. Available: https://cloud.google.com/docs

[65] Ian Goodfellow, Yoshua Bengio, and Aaron Courville, *Deep Learning*. MIT Press, 2016. Accessed: Apr. 04, 2025. [Online]. Available: https://www.deeplearningbook.org/

[66] M. K. Raynolds *et al.*, "A raster version of the Circumpolar Arctic Vegetation Map (CAVM)," *Remote Sens. Environ.*, vol. 232, p. 111297, Oct. 2019, doi: 10.1016/j.rse.2019.111297.

[67] J. L. Brown, O. J. F. Jr, J. A. Heginbottom, and E. S. Melnikov, "Circum-Arctic map of permafrost and ground-ice conditions," U.S. Geological Survey, 45, 1997. doi: 10.3133/cp45.

[68] G. Ghiasi *et al.*, "Simple Copy-Paste is a Strong Data Augmentation Method for Instance Segmentation," Jun. 23, 2021, *arXiv*: arXiv:2012.07177. doi: 10.48550/arXiv.2012.07177.

[69] J. Long, E. Shelhamer, and T. Darrell, "Fully convolutional networks for semantic segmentation," in *2015 IEEE Conference on Computer Vision and Pattern Recognition (CVPR)*, Jun. 2015, pp. 3431–3440. doi: 10.1109/CVPR.2015.7298965.

[70] T.-Y. Lin *et al.*, "Microsoft COCO: Common Objects in Context," in *Computer Vision – ECCV 2014*, D. Fleet, T. Pajdla, B. Schiele, and T. Tuytelaars, Eds., Cham: Springer International Publishing, 2014, pp. 740–755. doi: 10.1007/978-3-319-10602-1_48.

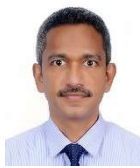
**Amal S. Perera.** (Member, IEEE), is a Senior Lecturer in Computer Science at the Department of Computer Science and Engineering, University of Moratuwa, Sri Lanka. He is currently working as a Data Scientist at the University of Connecticut as an invited post-doctoral research scholar. He has multiple research publications and a co-owner of a US patent. He won the ACM KDD Cup for data mining in 2006. He is an IEEE member since 2014. He has a PhD in Computer Science. His general areas of interest include data science, data mining, database systems, and software engineering. Currently he is working on developing and deploying deep learning-based computer vision models on Big Data on high performance computing environments.

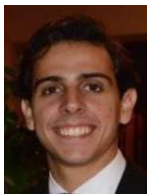
**David J. Fernandez** received a Bachelor's degree in Computer Science from the Georgia Institute of Technology in 2020 and a MSc in Computer Science with a concentration on computational perception and robotics from the Georgia Institute of Technology in 2021. During his graduate studies, David focused on embodied AI specifically in reinforcement learning and inverse reinforcement learning in order to solve complex tasks using robotic systems. After graduating, David joined Google as a Software Engineer at the Physical Automation team where he focuses on enabling robotics and automation solutions to help scale Google's growing datacenter fleet. David is an avid lover of the outdoors and is interested in leveraging advanced computing techniques for the betterment of the environment..

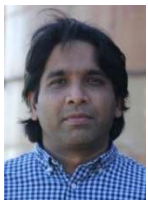
**Chandi Witharana** is a Professor at the University of Connecticut, with PhD in Remote Sensing. His research efforts broadly capture the methodological developments and adaptations to unseal faster, deeper, and more accurate analysis of large volumes of high-resolution remote sensing data. He conducts interdisciplinary remote sensing research with high international visibility, speaking equally to the transformational uses of remote sensing in environmental, industrial, and agricultural applications.

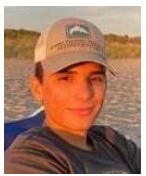
**Elias Manos** is a PhD student in Remote Sensing at the Department of Natural Resources, University of Connecticut. He is a geospatial deep learning practitioner with a background in GIS/remote sensing. He is a team member of the Permafrost Discovery Gateway project (funded by the NSF's Navigating the New Arctic program), His research centers on advancing remote sensing of Arctic permafrost regions by developing deep learning approaches to map the natural and built environments from high-resolution satellite imagery with unprecedented spatial and thematic detail. He is developing a deep learning approach to map Arctic infrastructure at a <1 m spatial resolution and circumpolar scale from Maxar satellite imagery to enable accurate assessment of infrastructure damage risk in the face of climate change-induced permafrost degradation.

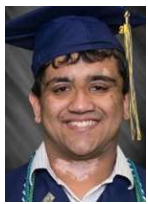
**Michael Pimenta** is a MS student in Remote Sensing at the Department of Natural Resources, University of Connecticut. He is a geospatial deep learning practitioner with a background in GIS/remote sensing. He is a team member of the Permafrost Discovery Gateway project (funded by the NSF's Navigating the New Arctic program), His research centers on advancing remote sensing of Arctic permafrost regions by developing deep learning approaches to map the trough network from high-resolution satellite imagery. to enable accurate estimates of the trough network.

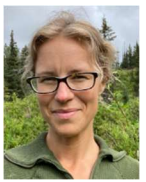
**Anna K. Liljedahl** received her Ph.D. in Hydrology from the University of Alaska Fairbanks in 2011. She is currently an Associate Scientist at the Woodwell Climate Research, and the project lead of the Permafrost Discovery Gateway team. Her work is rooted in a life-long love of water and a deep sense of connection to Alaska's environment and communities. Her research focuses on how climate change is altering the storage and movement of water in Arctic ecosystems. She has probed the impact of glacial melt and permafrost thawing on natural processes, like ponding and runoff, as well as built infrastructure, including hydropower. Dr. Liljedahl's work combines field measurements with remote sensing data to produce high-resolution computer models at watershed scales. Through projects like the Permafrost Discovery Gateway, Dr. Liljedahl strives to expand access to information and to expedite knowledge-generation from big data to serve earth scientists and communities facing climate impacts

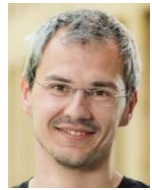
**Ingmar Nitze** received the bachelor's degree (B.Sc.) in geography from the Free University (FU) of Berlin, Berlin, Germany, in 2009, and the master's degree (M.Sc.) in geoinformation and visualization and the Doctorate degree in remote sensing (Ph.D./Dr.) from the University of Potsdam, Potsdam, Germany, in 2012 and 2018, respectively. From 2012 to 2014, he worked as a Research Assistant at the University College Cork, Cork, Ireland. He is currently a Researcher with the Permafrost Research Section, Alfred Wegener Institute (AWI) Helmholtz Centre for Polar and Marine Research, Potsdam. His research focuses on the detection and quantification of landscape dynamics in the circum-Arctic permafrost region using remote sensing and machine learning. Alfred Wegener Institute Helmholtz Centre for Polar and Marine Research, Potsdam, Germany

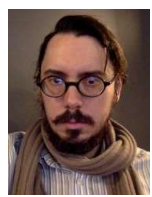
**Todd Nicholson** is a Research Programmer at NCSA (National Center for Supercomputing Applications) University of Illinoi at University of Illinois at Urbana-Champaign. His areas of interests include Graph theory, data science and machine learning, promoting software writing to community college and K-12 students. He obtaines his BSc and MSc from University of Illinoi at University of Illinois at Urbana-Champaign

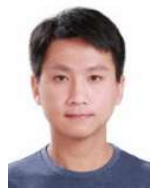
**Chia-Yu Hsu** received a Master's degree in Computer Science from Arizona State University in 2018. He is currently a research professional at Arizona State University. His research interests focus on applying machine learning and artificial intelligence techniques to address geographical big data challenges. In recent years, his work has emphasized GeoAI, including developing foundation models for Earth observation, advancing Arctic science with AI, and enhancing explainability in deep learning for geospatial applications.

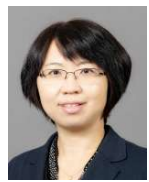
**Wenwen Li** received her Ph.D. in Earth System and Geoinformation Science from George Mason University in 2019. She is currently a Professor of GIScience in the School of Geographical Sciences and Urban Planning at Arizona State University, where she also directs the Spatial Analysis Research Center and the Cyberinfrastructure and Computational Intelligence Lab. Her research interests include cyberinfrastructure, geospatial big data, GeoAI, and their applications in data-intensive environmental and social sciences.

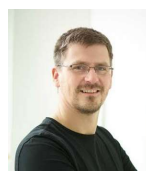
**Guido Grosse** received the Diploma (M.Sc.) degree in geology from the Technical University and Mining Academy Freiberg (TUBA), Freiberg, Germany, in 2021, and the Dr. rer. nat. (Ph.D.) degree in geosciences from the University of Potsdam, Potsdam, Germany, in 2005. He became a Post-Doctoral Researcher and then a Research Assistant Professor at the Geophysical Institute, University of Alaska Fairbanks, Fairbanks, AK, USA, in 2006 and 2009, respectively. He returned to Germany at the Alfred Wegener Institute (AWI) Helmholtz Centre for Polar and Marine Research, Potsdam and became a Full Professor on Permafrost in the Earth System jointly appointed by AWI and the University of Potsdam. Since 2016, he has been the Head of the Permafrost Research Section, AWI. His team increasingly develops and applies computer vision, machine learning, and deep learning methods in remote sensing of Arctic permafrost. He has authored more than 195 peer-reviewed publications, participated in more than 35 arctic expeditions and is involved in multiple international permafrost-related networks and research projects. His research focuses on remote sensing of landscape dynamics across broad spatial and temporal scales, hydrology, carbon cycling, and the impacts of climate change in Arctic permafrost regions. Dr. Grosse won an ERC Starting Grant in 2013